%% file: acl_latex.tex
\pdfoutput=1

\documentclass[11pt]{article}

\usepackage[preprint]{acl}

\usepackage{times}
\usepackage{latexsym}

\usepackage[T1]{fontenc}

\usepackage[utf8]{inputenc}

\usepackage{microtype}

\usepackage{inconsolata}

\usepackage{graphicx}
\usepackage{textcomp}
\usepackage{enumitem}
\usepackage{amsmath}
\usepackage{booktabs}
\usepackage{tcolorbox}
\usepackage{subcaption}
\usepackage{multirow} 
\usepackage{amssymb}
\usepackage[normalem]{ulem}
\usepackage{textcomp}
\newcommand{\xmltagfont}{\ttfamily}
\usepackage{enumitem}
\usepackage{tabularx}    
\usepackage{array}       

\usepackage{arydshln}     

\usepackage{xcolor} 
\definecolor{deepgreen}{rgb}{0,0.3,0}

\definecolor{planColor}{RGB}{255,127,0}     
\definecolor{searchColor}{RGB}{138,43,226}  
\definecolor{infoColor}{RGB}{34,139,34}     
\definecolor{thinkColor}{RGB}{0,128,128}    
\definecolor{answerColor}{RGB}{199,21,133}  

\usepackage{pifont}

\NewDocumentCommand{\heng}
{ mO{} }{\textcolor{red}{\textsuperscript{\textit{Heng}}\textsf{\textbf{\small[#1]}}}}

\NewDocumentCommand{\cheng}
{ mO{} }{\textcolor{orange}{\textsuperscript{\textit{Cheng}}\textsf{\textbf{\small[#1]}}}}

\NewDocumentCommand{\xiusi}
{ mO{} }{\textcolor{cyan}{\textsuperscript{\textit{Xiusi}}\textsf{\textbf{\small[#1]}}}}

\NewDocumentCommand{\qi}
{ mO{} }{\textcolor{deepgreen}{\textsuperscript{\textit{Qi}}\textsf{\textbf{\small[#1]}}}}

\NewDocumentCommand{\yi}
{ mO{} }{\textcolor{gray}{\textsuperscript{\textit{Yi}}\textsf{\textbf{\small[#1]}}}}

\title{Veri-R1: Toward Precise and Faithful Claim Verification via Online Reinforcement Learning}

\author{
Qi He$^{1,2}$, Cheng Qian$^{1}$, Xiusi Chen$^{1}$, Bingxiang He$^{3}$, Yi R. (May) Fung$^{1,4}$, Heng Ji$^{1}$\\
$^{1}$ University of Illinois Urbana-Champaign, 
$^{2}$ Fudan University \\
$^{3}$ Tsinghua University, 
$^{4}$ Hong Kong University of Science and Technology \\
\texttt{qhe22@m.fudan.edu.cn}, \texttt{\{xiusic,hengji\}@illinois.edu}
}

\begin{document}
\maketitle

\input{sections/0_abstract}
\input{sections/1_introduction}
\input{sections/2_related_work}

\input{sections/3_method}
\input{sections/4_experiment}
\input{sections/5_analysis}

\input{sections/6_conclusion}

\bibliography{custom}

\input{sections/7_appendix}

\end{document}

%% file: sections/0_abstract.tex
\begin{abstract}
Claim verification with large language models (LLMs) has recently attracted growing attention, due to their strong reasoning capabilities and transparent verification processes compared to traditional answer-only judgments. However, existing approaches to online claim verification, which requires iterative evidence retrieval and reasoning, still mainly rely on prompt engineering or pre-designed reasoning workflows, without unified training to improve necessary skills. Therefore, we introduce \textbf{Veri-R1}, an online reinforcement learning (RL) framework that enables an LLM to interact with a search engine and to receive reward signals that explicitly shape its planning, retrieval, and reasoning behaviors. This dynamic interaction of LLM with retrieval systems more accurately reflects real-world verification scenarios and fosters comprehensive verification skills. Empirical results show that Veri-R1 improves joint accuracy by up to 30\% and doubles the evidence score, often surpassing its larger-scale model counterparts. Ablation studies further reveal the impact of reward components, and the link between output logits and label accuracy. Our results highlight the effectiveness of online RL for precise and faithful claim verification, providing an important foundation for future research. We release our code to support community progress in LLM empowered claim verification.\footnote{The code, models and datasets are available at \url{https://github.com/H0key-22/Veri-R1}.}
\end{abstract}

%% file: sections/1_introduction.tex
\section{Introduction}
Claim verification with LLMs has emerged as an increasingly important and complex challenge in natural language processing. With the society producing an enormous number of claims on the internet and increasing LLM generated contents, the proliferation of unverified claims on the web has accelerated at an unprecedented rate. Manual verification by humans is no longer a feasible solution given the overwhelming volume of information. Consequently, automatic and effective approaches to claim verification are becoming a more urgent need~\cite{dmonte2024claim}. 

\begin{figure}
    \centering
    \includegraphics[width=\linewidth]{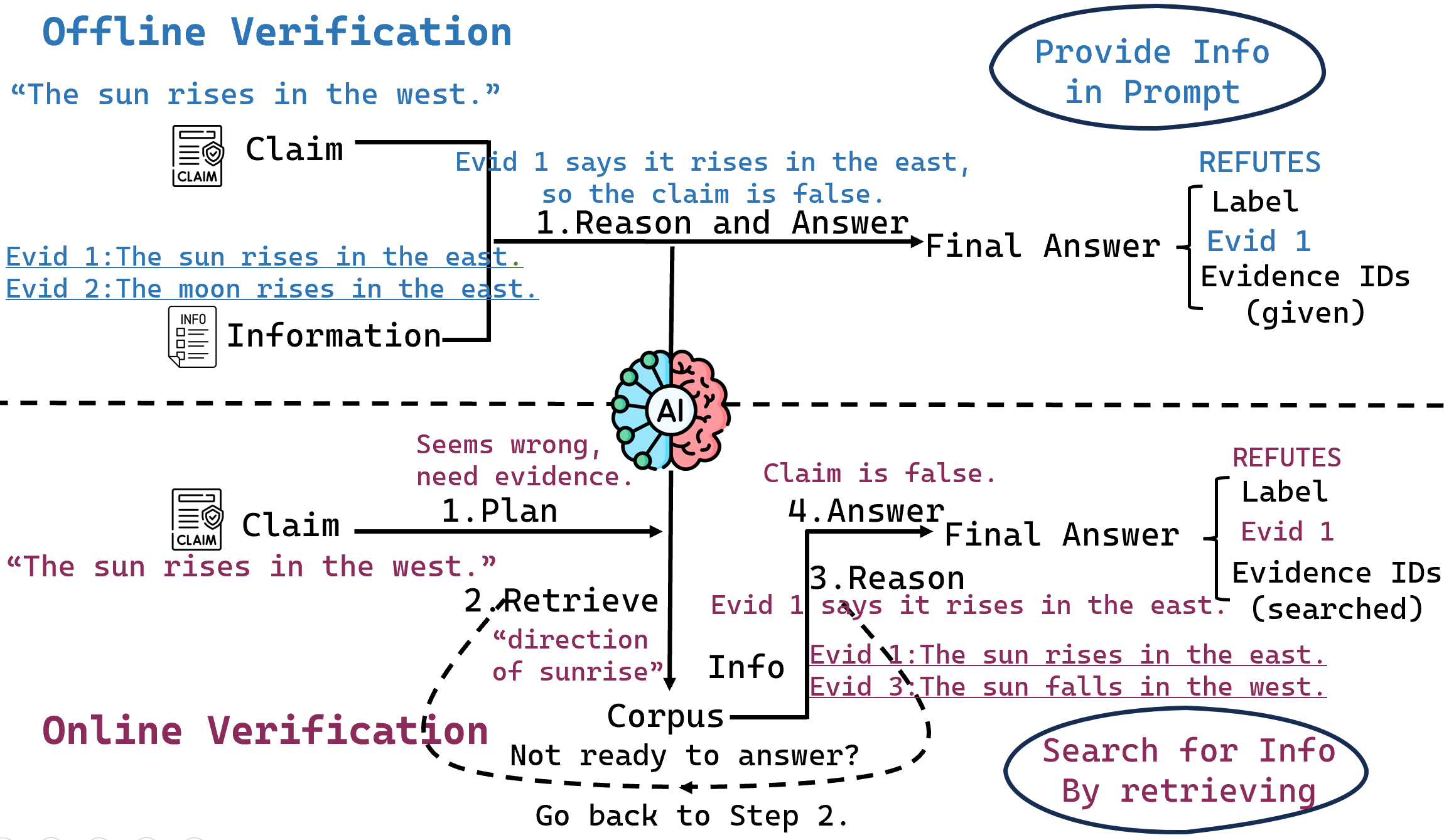}
    \caption{Conceptual comparison of \textbf{Offline Claim Verification} and \textbf{Online Claim Verification}. In the offline setting, models are provided with both the claim and relevant evidence, requiring only reasoning to produce the final answer. In the online setting, models must iteratively retrieve relevant information from a corpus before reasoning and producing the final answer}
    \label{fig:rollout}
\end{figure}

Previous studies~\cite{weng2022large, li2023self} have primarily focused on enhancing a model’s verification capability by providing both claims and corresponding evidence. This setting, referred to as Offline Claim Verification (OFFCV), as illustrated in Figure~\ref{fig:rollout}, requires the model to verify a given claim based solely on the provided evidence. However, in real-world scenarios, claim verification often lacks direct and concrete supporting evidence. Instead, the model must actively retrieve relevant information to assess the claim’s validity. We define this process as Online Claim Verification (ONCV), where the model is given only the claim and access to a trusted corpus. The model is asked to actively acquire and utilize relevant evidence to perform verification, as shown in Figure~\ref{fig:rollout}. In this work, we focus on the more realistic ONCV setting and evaluate model performance in this context.

Online claim verification poses additional great challenges to the LLMs since it demands all-around capabilities, which combine information retrieval, reasoning, and judgment. Similar to a sophisticated human claim verifier, the LLM must iteratively reason over retrieved content and interact with a search engine across multiple turns before producing a final judgment. Traditional approaches~\cite{hanselowski2018ukp, zhang2023relevance, pan2021zero} often adopt attribute-oriented methods~\cite{reddy2021newsclaims, reddy2022zero, li2022covid} to analyze claims, followed by a verifier that outputs the final decision without explanation~\cite{soleimani2020bert}. While LLMs have made the verification process more transparent~\cite{lee2020language, guan2023language}, they still exhibit notable limitations in handling the full ONCV pipeline. Recent work~\cite{trinh2025towards, kim2024can} has attempted to improve LLM performance on specific claim datasets by providing structured reasoning paths or introducing tailored analytical methods. However, these methods are typically specific to certain domains and challenges, whereas claim verification spans a wide range of domains~\cite{augenstein2019multifc, wang2024piecing} and encompasses diverse challenges such as multi-hop reasoning, entity disambiguation, numerical reasoning and more. Consequently, improving model performance in a more general claim verification setting requires addressing a central question:

\begin{center}
\textbf{\textit{How can we comprehensively enhance a model’s ability to search, reason, and judge across varied claim verification scenarios and challenges?}}
\end{center}

In this work, we propose \textbf{Veri-R1}, a novel training framework designed to enhance the online claim verification capability of LLMs within a unified pipeline. Unlike common approaches such as supervised fine-tuning (SFT)~\cite{zheng2025reasoning}, we adopt a reinforcement learning (RL) paradigm, as it has demonstrated superior generalization ability in numerous prior studies and does not require explicit reasoning trajectories for training. Moreover, we employ online RL as our primary methodology due to its previous empirical success~\cite{wang2025ragen} and its closer correspondence to real-world settings in which evidence should be retrieved and identified. During training rollouts, LLMs are required to iteratively search and reason across multiple turns before producing final answers. Since high-quality data is essential for effective training, and mislabeled samples can hinder the learning process, we filter and select samples from two high-quality datasets—FEVEROUS~\cite{aly2021feverous} and EX-FEVER~\cite{ma2023ex}—covering a wide range of claim verification challenges. To guide the training process, we design a task-specific reward function tailored to claim verification. Leveraging the ground-truth labels, the reward encourages models to learn robust judgment skills, while golden evidence is incorporated to ensure that models retrieve and identify evidence both completely and precisely. Furthermore, we conduct comprehensive comparisons among different training paradigms, including SFT and RL in both online and offline settings, and perform an ablation study to analyze the contribution of each reward component.

Empirically, online RL model from our Veri-R1 pipeline yielded the best performance in most cases compared to other training methods, yielding up to a 30\% absolute gain in joint accuracy, a 23\% improvement in verification accuracy, and a 22\% improvement in label accuracy on evaluation data. It also enhanced the evidence‐scoring metric by up to 150\%. Our ablation study shows that the evidence reward effectively improves the model’s ability to identify gold evidence, while the validity weight maintains training consistency by preventing reward hacking and encouraging correct label prediction with sufficient supporting evidence. Finally, we probe the relationship between model confidence and prediction correctness, finding that low confidence consistently correlates with low accuracy for SUPPORT/REFUTE labels. Moreover, larger models tend to avoid answering with NOT ENOUGH INFO, as they are more confident in their predictions.

In summary, our core contributions are as follows:
\begin{itemize}[topsep=0pt,noitemsep,leftmargin=*]
\item \textbf{Veri-R1 framework}: We propose a unified RL-based pipeline tailored for Online Claim Verification. By contrasting Online RL and Offline RL against SFT, we highlight how interactive, feedback-driven learning better captures the dynamics of real-world verification tasks.
\item \textbf{Data \& Reward Design}: We construct a high-quality dataset from FEVEROUS and EX-FEVER and design a robust reward system that jointly emphasizes multi-level accuracy and precise evidence retrieval and identification. This alignment of supervision with verification objectives ensures that models learn to reason in ways directly tied to label and evidence precision, while maintaining faithful reasoning trajectories.
\item \textbf{Confidence–Accuracy Analysis}: We conduct a detailed investigation of model confidence, showing that low-confidence predictions frequently correspond to errors in both SUPPORT and REFUTE cases. Moreover, we find that larger models tend to exhibit overconfidence, favoring definitive judgments while underutilizing the NOT ENOUGH INFO category. These results highlight an alternative avenue for supervising model outputs, while also underscoring that the overconfidence of larger-scale models constitutes a potential drawback.
\end{itemize}

%% file: sections/2_related_work.tex
\section{Related Work}

\subsection{LLM Empowered Claim Verification}
Early investigations have primarily focused on offline claim verification, leveraging LLMs as the main reasoner and verifier. For example, pioneering work \cite{buchholz2023assessing} demonstrated that LLMs can assess and validate factual assertions with promising accuracy. To further improve claim verification performance, several studies \cite{pisarevskaya2025zero, gong2025strive} have introduced structured reasoning prompts\cite{wei2022chain}, guiding LLMs to decompose claims into sequences of analytic steps \cite{vladika2025step, hu2024decomposition}, thereby enhancing transparency and fostering trust in their judgments. Other research has explored analytical tool integration, such as incorporating entity graphs to enforce systematic reasoning trajectories \cite{jeon2025graphcheck, huang2025graph}.

More recent work has shifted toward online claim verification, where evidence must be retrieved by the model. Retrieval-augmented frameworks \cite{vykopal2025generative, hagstrom2024reality} enable models to query external knowledge bases and rapidly incorporate emerging information. Given the additional challenges in the online setting, researchers have proposed new approaches—such as multi-agent systems\cite{hu2025coordinating} and program-guided reasoning\cite{pan2023fact,hu2025boost}—to help models adapt to diverse verification tasks.

Nevertheless, improving model performance for online claim verification across diverse challenges and domains remains an open problem. In this work, we introduce a framework designed to comprehensively enhance model verification capabilities in this complex setting.

\subsection{Online Reinforcement Learning}
Reinforcement learning (RL) \cite{kaelbling1996reinforcement} has long been recognized as a promising paradigm for enabling agents to interact with their environment and learn from feedback \cite{watkins1992q, rummery1994line}. Recently, RL techniques have emerged as a powerful mechanism for enhancing the inferential capabilities of large language models (LLMs) across a wide range of tasks, including mathematical problem solving \cite{shao2024deepseekmath,huang2025adactrladaptivecontrollablereasoning}, medical diagnosis \cite{lai2025med}, role-playing \cite{mou2024individual}, and dialogue generation \cite{chen2025rm}.

As the complexity of interactions between LLMs and their counterparts—such as users or external tools—increases~\cite{feng2025retool}, there is a growing need for online RL training frameworks that support multi-turn interactions. \citet{jin2025search} proposed an RL framework enabling models to interactively search for information, demonstrating the effectiveness of online RL training. Subsequently, several studies \cite{mei20252, xue2025simpletir} have sought to adapt the online RL pipeline to diverse application scenarios and improve training efficiency, for instance by refining rewards at each interaction turn \cite{wang2025stepsearch, zeng2025reinforcing}.

However, the effectiveness of online RL remains largely unexplored in the context of claim verification, and no prior work has systematically compared online RL with offline RL~\cite{levine2020offline}. In this work, we implement both RL paradigms for claim verification and conduct an extensive comparison to evaluate their respective advantages and limitations.

%% file: sections/3_method.tex
\section{Method}
\begin{figure*}
    \centering
    \includegraphics[width=\linewidth]{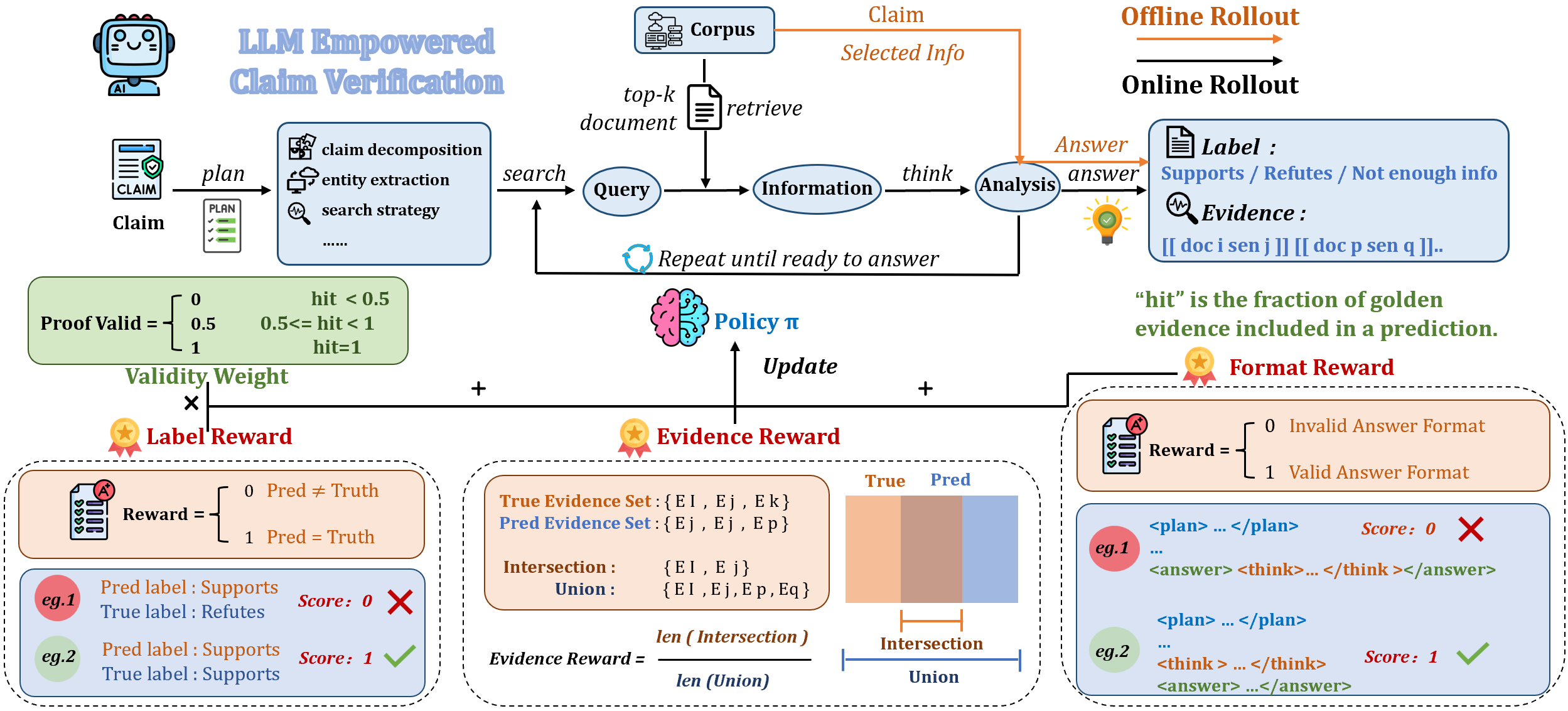}
    \caption{Comprehensive framework of Veri-R1, depicting the Online Claim Verification (ONCV) and Offline Claim Verification (OFFCV) workflows together with the calculation of label, evidence, and format rewards.}
    \label{fig:enter-label}
\end{figure*}

In the claim verification task, a model must not only determine the veracity of a claim but also provide evidences. Although supervised fine-tuning has been shown to enhance label prediction accuracy~\cite{zheng2025reasoning}, it typically requires high-quality reasoning trajectories and often falls short in terms of robustness and adaptability, particularly in complex or domain-specific settings. Reinforcement learning (RL) has demonstrated substantial promise in optimizing model behavior across a wide range of tasks and scenarios. Empirical evidence~\cite{qian2025toolrl} indicates that RL training can even surpass prevailing supervised fine‑tuning in terms of robustness and adaptability. Inspired by the power of RL training, we construct the \textbf{Veri-R1} training pipeline, built primarily on Online RL. Crucially, we devise a carefully calibrated reward function tailored to the claim‑verification task, which ensures that the model improves in both label accuracy and evidence accuracy.

\subsection{Task Definition}
Online Claim Verification (ONCV) denotes an end‑to‑end process in which a LLM autonomously leverages external search engines to gather evidence, performs intermediate reasoning steps, and produces a final veracity label. A typical verification trajectory comprises an initial planning stage, multiple search–reasoning iterations, and a concluding judgment accompanied by selected evidence. Concretely, given a claim $c$, the model first generates a verification plan $p$. At each iteration $i \in \{1, \dots, k\}$, it issues a search action $s_i$ with query $q_i$, retrieves information $i_i$ from the corpus, and executes an internal reasoning step $t_i$. Finally, the model produces an answer $a$, as detailed in Section~\ref{para:answer-format}. We denote the full trajectory as
\[
T = \bigl(p,\,(s_1,i_1,t_1),\,(s_2,i_2,t_2),\dots,(s_k,i_k,t_k),\,a \bigr).
\]

Under this framework, we optimize the model’s policy $\pi_\theta$ by maximizing the expected reward
\[
\max_\theta \; E_{T \sim \pi_\theta}\bigl[R(a)\bigr]
\]
where $R(a)$, described in Section~\ref{subsec:reward-design}, is a scalar feedback signal computed from the final answer.
In practice, we adopt \emph{Group Relative Policy Optimization} (GRPO)~\cite{shao2024deepseekmath}, where rewards are normalized within each group before policy updates. The GRPO objective can be formulated as
\begin{equation*}
\scalebox{0.75}{$
L(\theta) = \mathbb{E}\Bigl[
   \min\bigl(r(a)\,\hat R_{\mathrm{grp}}(a), \;
   \mathrm{clip}(r(a), 1-\epsilon, 1+\epsilon)\,\hat R_{\mathrm{grp}}(a)\bigr)
\Bigr],
$}
\end{equation*}
where 
\[
r(a) = \frac{\pi_\theta(a \mid s)}{\pi_{\theta_{\text{old}}}(a \mid s)}, 
\qquad 
\hat{R}_{\text{grp}}(a) = \frac{R(a) - \mu_g}{\sigma_g + \epsilon},
\]
with $\mu_g, \sigma_g$ denoting the mean and standard deviation of rewards within the group, respectively.

\subsection{Rollout Settings}

\paragraph{Offline Rollout}
Offline rollout refers to the process in which, given an initial prompt containing a target claim, supporting or refuting evidence, and contextual sentences, a model internally reasons based on these inputs—without any further interaction—and then generates its final verdict.

\paragraph{Online Rollout}
By contrast, online rollout mandates adherence to the specified algorithmic trajectory: the model may issue up to \(k\) search turns, interleaving retrieval and reasoning to support or refute the claim.

\paragraph{Token Delimiters}
To enforce procedural rigor and faclitate answer parsing, each component of the workflow must be delimited by designated tokens:

\noindent
\texttt{\color{planColor}\xmltagfont <plan>…</plan>}: Strategic planning\\[0.8ex]
\texttt{\color{searchColor}\xmltagfont <search>…</search>}: Retrieval queries\\[0.8ex]
\texttt{\color{thinkColor}\xmltagfont <think>…</think>}: Reasoning Process\\[0.8ex]
\texttt{\color{answerColor}\xmltagfont <answer>…</answer>}: Final judgment

\paragraph{Answer Format}\label{para:answer-format}
The model’s output must consist solely of two elements within the pre-defined tag: \\
\texttt{\color{answerColor}\xmltagfont <answer>} \\
\textbf{Label}: one of \texttt{SUPPORT}, \texttt{REFUTE}, or \texttt{NOT ENOUGH INFO} \\
\textbf{Evidence}: a list of evidence id formatted as \\$[[\texttt{evid\_id}_1],\ [\texttt{evid\_id}_2]],\dots$ \\
\texttt{\color{answerColor}\xmltagfont </answer>} \\
We adopt a strict answer format to simplify parsing, using \texttt{[[]]} to delimit each evidence ID and guarantee accurate extraction.

\subsection{Reward Design}\label{subsec:reward-design}
Rule‐oriented reward strategies have demonstrated strong efficacy across a wide range of experiments and gain broad support~\cite{icarte2022reward,li2025torl,wang2025otc}. Because an effective policy often needs to balance multiple objectives, many studies combine several sub-reward functions using a linear weighted sum. In our framework, we define two primary components—\textbf{label reward} and \textbf{evidence reward}—to more directly steer the verification process. Additionally, we incorporate a \textbf{format reward} to enforce conformity to the desired output schema and to mitigate generation errors during reasoning. Finally, we introduce a validity weight to further steer and stabilize the training process.

\paragraph{Format Reward}
We enforce format compliance according to the following principles: (1) \textbf{Tag adherence}: all actions must strictly follow the prompt and be emitted within the prescribed tags; (2) \textbf{No extraneous tags}: the model must not introduce \verb|<information>…</information>| (or any other undeclared tags) on its own. Invalid \verb|<information>| tags are detected based on whether they immediately follow a \verb|<search>| tag, since the system automatically identifies \verb|<search>| tags and appends the corresponding \verb|<information>| tags.\\
Finally, if the verification transcript conforms to all of these rules, it is awarded a format reward of~1; otherwise, it receives~0:
\[
R_{\mathrm{format}} =
\begin{cases}
1, & \text{if all format rules are satisfied}, \\
0, & \text{otherwise}.
\end{cases}
\]

\paragraph{Evidence Reward}
Evidence is essential for validating a model’s prediction and guarding against correct guesses made by chance. Accordingly, we define the evidence reward as the ratio of the intersection to the union between the predicted evidence set and the gold‐standard evidence set. This reward is maximized only when the model retrieves all true evidence while avoiding irrelevant selections.
\[
R_{\mathrm{evidence}}
= \frac{\bigl|E_{\mathrm{pred}}\cap E_{\mathrm{gold}}\bigr|}
       {\bigl|E_{\mathrm{pred}}\cup E_{\mathrm{gold}}\bigr|},
\quad R_{\mathrm{evidence}} \in [0,1].
\]

\noindent \(E_{\mathrm{pred}}\) denotes the set of evidence items selected by the model, and \(E_{\mathrm{gold}}\) denotes the set of ground‐truth evidence items.

\paragraph{Label Reward}
In claim verification, the label space is limited to \texttt{SUPPORT}, \texttt{REFUTE} or \texttt{NOT ENOUGH INFO}. Because label accuracy is the most critical metric for developing reliable verification models, we amplify the base label reward by giving it higher value, thereby boosting its relative weight in the overall reward function.
\[
R_{\mathrm{label}} =
\begin{cases}
2, & \hat{y} = y,\\
0, & \text{otherwise},
\end{cases}
\]
\(\hat{y}\) is the model’s predicted label and \(y\) is the ground‐truth label.

\paragraph{Validity Weight}
There are cases in which a model arrives at the correct label via an unsound or “shortcut” reasoning path—for instance, predicting \texttt{SUPPORT} after verifying only a single subclaim. Such behavior undermines the model’s capacity for genuine scrutiny and may even lead it astray. To guard against this, we introduce a validity constraint on the label reward. We define the hit rate as
\[
h = \frac{\lvert E_{\mathrm{pred}}\cap E_{\mathrm{gold}}\rvert}{\lvert E_{\mathrm{gold}}\rvert}.
\]
A policy can earn the full label reward (\(w_{\mathrm{validity}} = 1\)) for a \texttt{SUPPORT} or \texttt{REFUTE} decision only if it “hits” all of the gold evidence. To mitigate sparsity, we grant a half reward (\(w_{\mathrm{validity}} = 0.5\)) whenever the hit rate exceeds 50\%. We set the threshold at 50\% since it is the natural midpoint of the evidence hit rate: retrieving more than half of the gold evidence indicates that the model has captured the majority of the reasoning path. For \texttt{NOT ENOUGH INFORMATION (NEI)} labels, however, we apply no validity weighting, since NEI cases often lack explicit evidence or involve only partially relevant facts.
\vspace{-0.3em}
\[
w_{\mathrm{validity}} =
\begin{cases}
1, & y = \mathrm{N} \lor \bigl(y\in\{\mathrm{S},\mathrm{R}\}\land h=1\bigr),\\[6pt]
0.5, & y\in\{\mathrm{S},\mathrm{R}\}\land h>0.5,\\[4pt]
0, & \text{otherwise}.
\end{cases}
\]
N, S, and R abbreviate \texttt{NEI}, \texttt{SUPPORT}, and \texttt{REFUTE}, respectively. 

\paragraph{Final Reward}
The reward components are subsequently integrated to formulate the final reward for optimization:
\[
R_{\mathrm{final}} = R_{\mathrm{label}} \cdot w_{\mathrm{validity}} + R_{\mathrm{evidence}} + R_{\mathrm{format}}
\]

%% file: sections/4_experiment.tex
\section{Experiment}
\subsection{Dataset}
To comprehensively train and evaluate the base model, we assembled a claim‑verification corpus encompassing diverse challenges to ensure robust, generalizable performance. Recognizing the crucial role of data quality, we also implemented a filtering pipeline to retain only high‑quality examples. Details of dataset are described in appendix~\ref{appendix:dataset}

\subsection*{Training Dataset}
\paragraph{FEVEROUS}~\cite{aly2021feverous} is a comprehensive verification dataset presenting diverse challenges—such as entity disambiguation, numerical reasoning, and more.
\paragraph{EX‑FEVER}~\cite{ma2023ex} is designed to evaluate the explainability of the verification process, comprising thousands of multi‑hop claims.

\subsection*{Hold-Out Dataset}
\paragraph{FEVER}~\cite{thorne2018fever} is a canonical claim‑verification benchmark composed largely of concise, single‑sentence claims.
\paragraph{HOVER}~\cite{jiang2020hover} defines a more challenging multi-hop verification task, demanding reasoning across 2 to 4 hops over Wikipedia entities.
\paragraph{SciFACT}~\cite{wadden2020fact} is a scientific-domain fact-checking corpus containing expert-annotated claims from research papers.
\subsection*{Data Filtering}
\begin{figure}
    \centering
    \includegraphics[width=\linewidth]{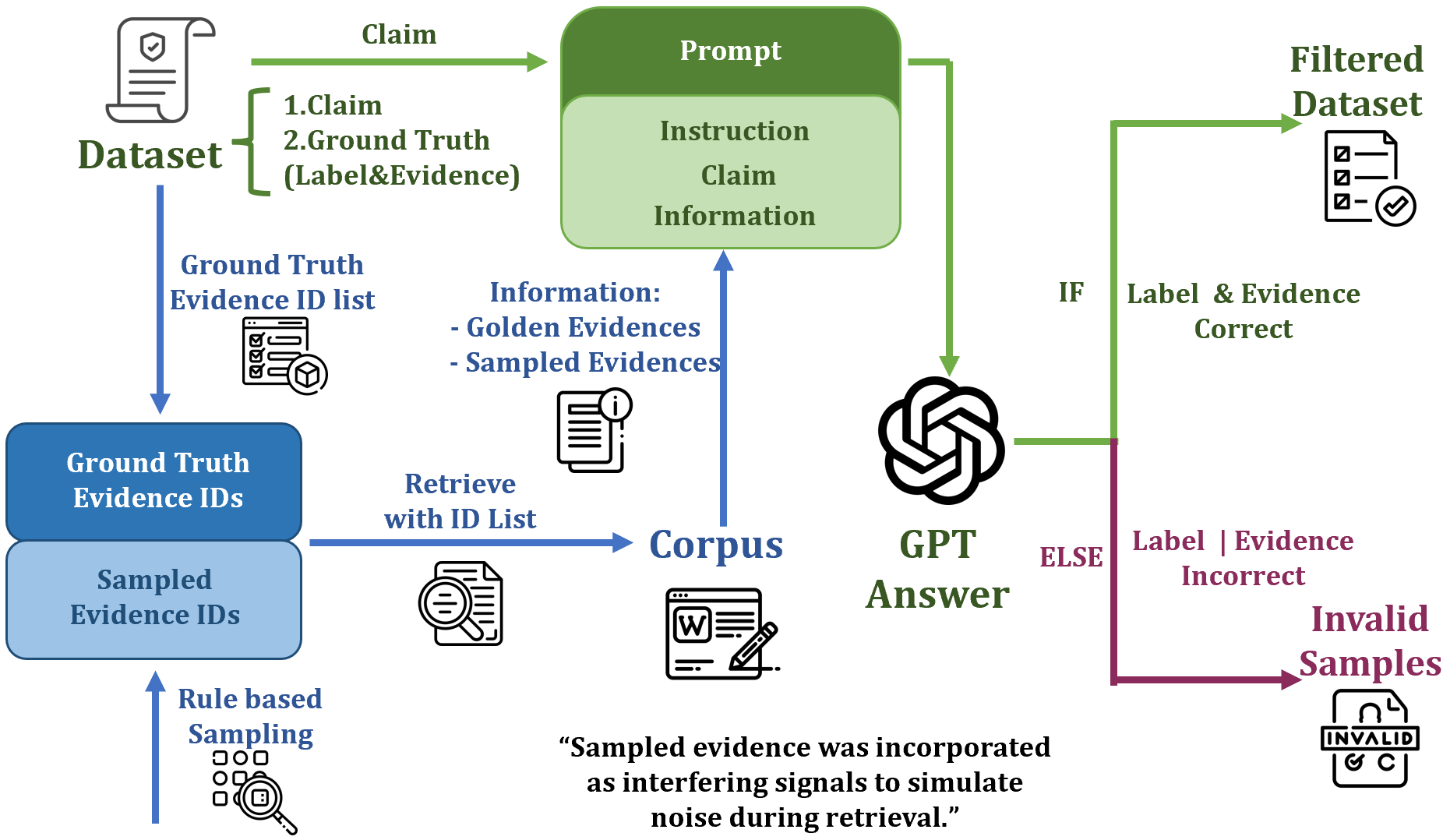}
    \caption{To mitigate annotation-related issues and ambiguities in the raw dataset, we developed a pipeline that simulates offline rollout, using GPT-4o to filter and preserve only high-quality data.}
    \label{fig:data_filter}
\end{figure}

Given the ambiguous nature of claims, there are often some ambiguity for a claim. We also find some toxic sample in the dataset may disturb a smooth training process. Considering the training effectiveness and evaluation credibility, we decided to filter the dataset with assistance of GPT-4o. We simulate the process of offline rollout shown in Figure~\ref{fig:data_filter}, where we provide the model with claims and sufficient information. As is depicted in the figure, we only select the sample where GPT-4o is totally correct, which means it predicts the right label and evidence with no deviation. In general, approximately 70\% of the samples are retained during the filtering process.
 
\subsection{Experiment Setting}
\input{tables/main_prompt}

\paragraph{Training}
Our online claim verification system builds upon Verl~\cite{sheng2025hybridflow} and Search‑R1~\cite{jin2025search}. During RL training, we employ the GRPO algorithm to preferentially learn from high‑quality trajectories. We cap the number of search turns at three and require the model to produce its answer once this budget is exhausted.

\paragraph{Evaluation}
We evaluate our approach across five datasets: the \textsc{FEVEROUS} and \textsc{EX-FEVER} test sets, as well as selected subsets of \textsc{FEVER}, \textsc{HOVER}, and \textsc{SciFACT}. Within each dataset, \textbf{the class labels are uniformly distributed} to eliminate potential biases arising from class imbalance. In addition to the evidence score (as described in our methodology), we report three distinct accuracy metrics:
\begin{itemize}[noitemsep,leftmargin=*]
    \item \textbf{Joint Accuracy (Joint Acc)}: the proportion of instances in which both the predicted label and the retrieved evidence are correct.
    \item \textbf{Verification Accuracy (Veri Acc)}: the proportion of instances in which the model predicts the correct label and retrieves all gold-standard evidence (allowing redundant evidence).
    \item \textbf{Label Accuracy (Label Acc)}: the proportion of instances in which the predicted class label is correct.
\end{itemize}
All evaluations are conducted in an \textbf{online setting}, where the model retrieves the necessary information from the corpus using queries. 
For instances classified as \textit{NOT ENOUGH INFO (NEI)}, we relax the requirement of retrieving all evidence items, since such cases often lack sufficient evidence or contain inherently subjective evidence annotations.
\input{tables/evaluation_all}
\input{tables/joint_acc}
\subsection{Baseline}
To evaluate the effectiveness of online RL training, we compare our approach against the following baselines:
\begin{enumerate}[leftmargin=*]
    \item \textbf{Raw Instruct Model}: the original instruction-tuned model without any additional training. For baselines, we adopt the widely used Qwen2.5‑3B‑Instruct~\cite{team2024qwen2} and Llama3.2‑3B‑Instruct models~\cite{dubey2024llama}.
    \item \textbf{SFT Model}: the instruct model supervised fine-tuned(SFT) on simulated reasoning paths generated by GPT-4o for samples in the training set. We prompt GPT-4o using the template in Figure~\ref{tab:Online_RL} to guide it in producing high-quality verification reasoning for each claim. We perform LoRA adaptation~\cite{hu2022lora} using the llama‑factory toolkit~\cite{zheng2024llamafactory}.
    \item \textbf{Offline RL Model}: the instruct model trained using offline rollouts based on the prompt described in the appendix. The training follows the same reward function outlined in Section~\ref{subsec:reward-design}. We select the checkpoint with the highest validation accuracy for evaluation.
    \item \textbf{Raw Instruct Model with Larger Scale}: we use the 7B variant of the Qwen Instruct model and the 8B variant of the Llama Instruct model to assess the effect of scale.
\end{enumerate}

\subsection{Result}
\subsection*{Main Result}

The comparison between online training and baseline methods on the training dataset's test set (FEVEROUS and EX-FEVER) and the held-out datasets (FEVER, HOVER, and SciFACT )is presented in Table~\ref{tab:results_all}. Table~\ref{tab:joint_acc_all} additionally reports a label-wise comparison of joint accuracy. We also record evidence scores of all models across the five datasets in Table~\ref{tab:evidence_scores}. From these results, we draw several key observations:
\input{tables/evidence_score}
\paragraph{Online claim verification remains a challenging task for current models.} Even the state-of-the-art model, GPT-4o, achieves joint accuracies ranging from 26.64\% to 59.30\% across different datasets. This highlights the inherent difficulty of online claim verification, which requires iterative reasoning, information retrieval, and judgment. Furthermore, models face additional challenges stemming from multi-hop reasoning, the need for induction, and the ambiguous nature of many claims.

\paragraph{Online RL training substantially enhances models’ verification capabilities.}Models trained with Online RL achieve the highest scores across all metrics among models of the same parameter scale and, in many cases, even outperform larger-scale counterparts. This outcome aligns with our expectations, as online training mirrors the evaluation environment, enabling the model to better adapt to the system. These findings underscore the effectiveness of our Veri-R1 framework.

\paragraph{RL training outperforms SFT in the claim verification domain.} In most cases, models trained with RL achieve higher scores than their SFT counterparts. A possible reason is that SFT relies on reasoning trajectories generated by GPT-4o, which may lead the model to imitate the format of reasoning rather than genuinely learning the verification process. Moreover, these trajectories may differ from the reasoning logic required for other datasets, thereby limiting the model’s generalizability. In contrast, RL enables the model to actively explore and learn from higher-quality trajectories, which substantially enhances its verification capability.

\subsection*{Other Findings}

In addition to demonstrating the efficacy of online reinforcement learning, we identify several other noteworthy findings, which are presented below.
\paragraph{Supervised Fine-Tuning does not guarantee improvement}
In many cases, SFT versions underperform their base counterparts, suggesting that pure supervised tuning can lead to overfitting on training data distributions. This overfitting may reduce the model’s ability to generalize to out-of-distribution claims, especially when real-world fact verification requires handling noisier or more varied evidence than the training set. Additionally, SFT may bias the model toward producing confident but incorrect predictions if the supervision data lacks sufficient coverage of borderline or ambiguous cases.

\paragraph{Model size advantage is task-dependent}
As is shown in Table~\ref{tab:joint_acc_all}, larger models (e.g., LLaMA3.1-8B) tend to excel in SUPPORT/REFUTE judgments, likely because increased parameter capacity enhances their ability to absorb broad factual knowledge, retrieve relevant information, and perform multi-hop reasoning when sufficient evidence exists. In contrast, NEI  detection requires a different skill set—careful uncertainty calibration, recognition of evidence gaps, and the ability to resist making unwarranted inferences. These abilities are less tied to sheer model size and more dependent on training signals that penalize overconfident answers in the absence of proof. Online RL appears to aid NEI performance by encouraging more cautious, evidence-driven predictions, as seen in cases like Qwen2.5-3B-Instruct-OnlineRL’s high NEI accuracy in EX-FEVER.

\paragraph{REFUTE is generally the hardest label for models}
Across datasets, REFUTE (or CONTRADICTS) accuracies are usually lower and more variable than SUPPORT or NEI. This difficulty likely stems from the need to retrieve not just relevant evidence, but specific counter-evidence that clearly contradicts the claim. Contradictions often require more precise reasoning about causal relationships, negation, or temporal mismatches, and errors in evidence retrieval or reasoning chains can easily lead the model to misclassify REFUTE as NEI or SUPPORT.

%% file: tables/main_prompt.tex
\definecolor{planColor}{RGB}{255,127,0}     
\definecolor{searchColor}{RGB}{138,43,226}   
\definecolor{infoColor}{RGB}{34,139,34}      
\definecolor{thinkColor}{RGB}{0,128,128}     
\definecolor{answerColor}{RGB}{199,21,133}   

\begin{figure*}[t]
\centering
\resizebox{0.85\textwidth}{!}{%
\begin{tcolorbox}[colback=yellow!5!white, colframe=green!75!black,
  title=Claim Verification Assistant Prompt, boxrule=0.3mm,
  width=\textwidth, arc=3mm, auto outer arc=true]
You are a claim-verification assistant. You MUST follow this protocol exactly: \\
[4pt]
\textcolor{planColor}{\texttt{<plan>…</plan>}} \\[-2pt]
– Once at the start: sketch your high-level strategy, such as claim decomposition, entity recognition, etc. \\
[4pt]
\textcolor{searchColor}{\texttt{<search>…</search>}} \\[-2pt]
– When you need a fact: emit exactly this tag with your query.\\
– To make the most of your search turns, don’t repeat identical queries.\\
– You can search at most three times. \\
[4pt]
\textcolor{infoColor}{\texttt{<information>}}\\
\quad\texttt{[[e\_1]]: info1}\\
\quad\texttt{[[e\_2]]: info2}\\
\quad\texttt{…}\\
\textcolor{infoColor}{\texttt{</information>}} \\[-2pt]
– You will be given claim related information in the format above. \\
[4pt]
\textcolor{thinkColor}{\texttt{<think>…</think>}} \\[-2pt]
– Use for every piece of reasoning; do not state your final verdict here.\\
– You must conduct reasoning inside \texttt{<think>} and \texttt{</think>} first every time you get new information. \\
[4pt]
\textcolor{answerColor}{\texttt{<answer>}}\\
\quad Label: SUPPORT / REFUTE / NOT ENOUGH INFO\\
\quad Evidence: [[e\_1]], [[e\_3]], …\\
\textcolor{answerColor}{\texttt{</answer>}} \\[-2pt]
– Emit exactly once at the end, no extra text or tags.\\
– Evidence ids such as \texttt{e\_1} will be replaced by real ids from the corpus. Include only those ids in your evidence list.\\
- Evidence outputs must strictly enforce the format [[e\_i]], [[e\_j]]…\\
- Answer Labels respectively stand for:\\
\quad SUPPORT: The claim is consistent with the cited evidence and the evidence is sufficient to confirm the claim.\\
\quad REFUTE: The claim contradicts the cited evidence and the evidence is sufficient to disprove the claim.\\
\quad NOT ENOUGH INFO: The available evidence is insufficient to determine whether the claim is true or false.\\
[4pt]
– Process: \textcolor{planColor}{\texttt{plan}} → (\textcolor{searchColor}{\texttt{search}} → \textcolor{infoColor}{\texttt{information}} → \textcolor{thinkColor}{\texttt{think}}) repeat until conclusion → \textcolor{answerColor}{\texttt{answer}} \\
[4pt]
Verify the claim: \texttt{\{claim\}}
\end{tcolorbox}%
}
\caption{System Prompt for Online Claim Verification.}
\label{tab:Online_RL}
\end{figure*}

%% file: tables/evaluation_all.tex
\begin{table*}[!t]
\centering
\resizebox{1\linewidth}{!}{
\begin{tabular}{l ccc ccc ccc ccc ccc}
\toprule
\textbf{Model} 
& \multicolumn{3}{c}{\textbf{FEVEROUS}} 
& \multicolumn{3}{c}{\textbf{EX-FEVER}} 
& \multicolumn{3}{c}{\textbf{FEVER}} 
& \multicolumn{3}{c}{\textbf{SciFACT}} 
& \multicolumn{3}{c}{\textbf{HOVER}} \\
 & \textbf{Joint Acc} & \textbf{Veri Acc} & \textbf{Label Acc} 
 & \textbf{Joint Acc} & \textbf{Veri Acc} & \textbf{Label Acc} 
 & \textbf{Joint Acc} & \textbf{Veri Acc} & \textbf{Label Acc} 
 & \textbf{Joint Acc} & \textbf{Veri Acc} & \textbf{Label Acc} 
 & \textbf{Joint Acc} & \textbf{Veri Acc} & \textbf{Label Acc} \\
\midrule
GPT-4o & 26.64\% & 40.70\% & 64.96\% & 24.71\% & 26.48\% & 54.75\% 
       & 41.11\% & 60.44\% & 74.33\% 
       & 38.26\% & 54.43\% & 74.96\% 
       & 59.30\% & 61.90\% & 73.60\% \\
\midrule
Qwen2.5-3B-Instruct              & 16.89\% & 24.15\% & 49.55\% & 17.62\% & 19.90\% & 42.71\% 
                                 & 19.00\% & 47.22\% & 62.78\% & 22.78\% & 37.27\% & 54.43\% 
                                 & 46.30\% & 50.50\% & 62.30\% \\
Qwen2.5-3B-Instruct-SFT          & 16.33\% & 23.81\% & 47.39\% & 17.62\% & 20.28\% & 42.71\% 
                                 & 19.89\% & 47.56\% & 62.00\% & 23.77\% & 36.85\% & 55.56\% 
                                 & 44.30\% & 49.10\% & 60.90\% \\
Qwen2.5-3B-Instruct-OfflineRL    & \underline{19.16\%} & \underline{27.89\%} & 53.40\% 
                                 & \underline{18.00\%} & \underline{20.41\%} & \underline{50.19\%} 
                                 & \underline{30.11\%} & \underline{51.22\%} & \underline{67.22\%} 
                                 & \underline{28.41\%} & \underline{39.94\%} & \underline{65.12\%} 
                                 & 45.80\% & 51.00\% & 61.50\% \\
Qwen2.5-7B-Instruct              & 14.17\% & 25.51\% & \underline{53.97\%} & 14.96\% & 17.62\% & 46.39\% 
                                 & 26.00\% & 49.78\% & 66.11\% & 24.33\% & 36.01\% & \textbf{65.96\%} 
                                 & \underline{53.60\%} & \textbf{57.10\%} & \textbf{65.70\%} \\
Qwen2.5-3B-Instruct-OnlineRL     & \textbf{28.91\%} & \textbf{36.28\%} & \textbf{61.22\%} 
                                 & \textbf{31.69\%} & \textbf{32.45\%} & \textbf{61.09\%} 
                                 & \textbf{49.11\%} & \textbf{55.89\%} & \textbf{69.56\%} 
                                 & \textbf{38.82\%} & \textbf{43.18\%} & 63.43\% 
                                 & \textbf{53.70\%} & \underline{55.10\%} & \underline{63.70\%} \\
\midrule
Llama3.2-3B-Instruct             & 13.27\% & 19.27\% & 41.38\% & 17.24\% & 20.15\% & 37.26\% 
                                 & 13.11\% & 37.67\% & 53.78\% & 14.35\% & 24.75\% & 47.54\% 
                                 & 31.70\% & 36.30\% & 56.00\% \\
Llama3.2-3B-Instruct-SFT         & 12.24\% & 18.48\% & 43.65\% & 15.08\% & 18.76\% & 37.39\% 
                                 & 16.00\% & 41.67\% & 56.00\% & 15.61\% & 26.44\% & 50.49\% 
                                 & 32.50\% & 38.40\% & 57.40\% \\
Llama3.2-3B-Instruct-OfflineRL   & 16.10\% & 21.88\% & 44.10\% & 17.49\% & 20.79\% & 41.83\% 
                                 & 17.89\% & 37.78\% & 57.67\% & \underline{18.85\%} & 24.61\% & 51.76\% 
                                 & 35.90\% & 40.60\% & 57.20\% \\
Llama3.1-8B-Instruct             & \textbf{19.73\%} & \textbf{27.32\%} & \underline{50.91\%} 
                                 & \underline{24.08\%} & \underline{26.87\%} & \underline{52.34\%} 
                                 & \underline{23.56\%} & \textbf{53.33\%} & \underline{67.44\%} 
                                 & 16.32\% & \underline{32.63\%} & \underline{58.93\%} 
                                 & \underline{51.50\%} & \textbf{55.60\%} & \textbf{67.70\%} \\
Llama3.2-3B-Instruct-OnlineRL    & \underline{19.27\%} & \underline{26.30\%} & \textbf{53.17\%} 
                                 & \textbf{28.52\%} & \textbf{30.29\%} & \textbf{59.32\%} 
                                 & \textbf{40.11\%} & \underline{53.11\%} & \textbf{68.44\%} 
                                 & \textbf{31.65\%} & \textbf{44.30\%} & \textbf{66.53\%} 
                                 & \textbf{52.60\%} & \underline{54.90\%} & \underline{65.40\%} \\
\bottomrule
\end{tabular}
}
\caption{Unified performance comparison across FEVEROUS, EX-FEVER, FEVER, SciFACT, and HOVER datasets. 
\textbf{Bold numbers} denote the best performance within each group (Qwen and Llama), 
and \underline{underlined numbers} denote the second-best performance.}
\label{tab:results_all}
\end{table*}

%% file: tables/joint_acc.tex
\begin{table*}[!t]
\centering
\resizebox{1\linewidth}{!}{
\begin{tabular}{l ccc ccc ccc ccc cc}
\toprule
\textbf{Model}
& \multicolumn{3}{c}{\textbf{FEVEROUS (Joint Acc)}} 
& \multicolumn{3}{c}{\textbf{EX-FEVER (Joint Acc)}} 
& \multicolumn{3}{c}{\textbf{FEVER (Joint Acc)}} 
& \multicolumn{3}{c}{\textbf{SciFACT (Joint Acc)}} 
& \multicolumn{2}{c}{\textbf{HOVER (Joint Acc)}} \\
& \textbf{SUPPORTS} & \textbf{REFUTES} & \textbf{NEI} 
& \textbf{SUPPORT} & \textbf{REFUTE} & \textbf{NEI} 
& \textbf{SUPPORTS} & \textbf{REFUTES} & \textbf{NEI} 
& \textbf{SUPPORT} & \textbf{CONTRADICT} & \textbf{NEI} 
& \textbf{SUPPORTED} & \textbf{NOT SUPPORTED} \\
\midrule
GPT-4o 
& 32.65\% & 4.08\% & 43.19\% 
& 16.73\% & 13.31\% & 44.11\% 
& 44.00\% & 38.00\% & 41.33\% 
& 16.46\% & 18.99\% & 79.32\% 
& 50.60\% & 68.00\% \\
\midrule
Qwen2.5-3B-Instruct 
& 10.20\% & 0.00\% & 40.48\% 
& 6.46\% & 4.56\% & \underline{41.83\%} 
& 5.00\% & 1.67\% & \underline{50.33\%} 
& 2.11\% & 2.53\% & 63.71\% 
& 27.00\% & \underline{65.60\%} \\
Qwen2.5-3B-Instruct-SFT 
& 7.82\% & 0.34\% & \underline{40.82\%} 
& 8.75\% & 3.80\% & 40.30\% 
& 7.00\% & 3.00\% & 49.67\% 
& 1.27\% & 2.11\% & 67.93\% 
& 24.00\% & 64.60\% \\
Qwen2.5-3B-Instruct-OfflineRL 
& \underline{19.73\%} & \underline{4.08\%} & 33.67\% 
& 7.98\% & 5.32\% & 40.68\% 
& 25.67\% & 21.00\% & 43.67\% 
& 5.91\% & \underline{10.13\%} & \underline{69.20\%} 
& 28.80\% & 62.80\% \\
Qwen2.5-7B-Instruct 
& 16.33\% & 2.72\% & 23.47\% 
& \underline{11.03\%} & \underline{6.84\%} & 27.00\% 
& \underline{30.67\%} & \underline{23.33\%} & 24.00\% 
& \underline{8.86\%} & \underline{10.13\%} & 54.01\%
& \underline{34.00\%} & \textbf{73.20\%} \\
Qwen2.5-3B-Instruct-OnlineRL 
& \textbf{30.27\%} & \textbf{11.22\%} & \textbf{45.24\%} 
& \textbf{17.49\%} & \textbf{10.65\%} & \textbf{66.92\%} 
& \textbf{45.67\%} & \textbf{46.67\%} & \textbf{55.00\%} 
& \textbf{11.39\%} & \textbf{14.77\%} & \textbf{90.30\%} 
& \textbf{42.00\%} & 65.40\% \\
\midrule
Llama3.2-3B-Instruct 
& 3.74\% & 0.00\% & \underline{36.05\%} 
& 7.22\% & 1.90\% & 42.59\% 
& 3.67\% & 3.33\% & 32.33\% 
& 0.42\% & 0.42\% & 42.19\% 
& 17.40\% & 46.00\% \\
Llama3.2-3B-Instruct-SFT 
& 5.10\% & 0.68\% & 30.95\% 
& 7.60\% & 1.90\% & 35.74\% 
& 4.33\% & 4.67\% & \textbf{39.00\%} 
& 0.42\% & 0.42\% & 45.99\% 
& 17.80\% & 47.20\% \\
Llama3.2-3B-Instruct-OfflineRL 
& 11.22\% & 2.04\% & 35.03\% 
& 8.75\% & \underline{9.13\%} & 34.60\% 
& 13.33\% & 13.33\% & 27.00\% 
& 1.27\% & 1.27\% & \underline{54.01\%} 
& 21.40\% & 50.40\% \\
Llama3.1-8B-Instruct 
& \underline{16.67\%} & \underline{3.40\%} & \textbf{39.12\%} 
& \underline{15.97\%} & 6.84\% & \underline{49.43\%} 
& \underline{23.00\%} & \underline{14.00\%} & 33.67\% 
& \underline{3.80\%} & \underline{2.11\%} & 43.04\%
& \underline{42.60\%} & \underline{60.40\%} \\
Llama3.2-3B-Instruct-OnlineRL 
& \textbf{21.77\%} & \textbf{4.08\%} & 31.97\% 
& \textbf{20.91\%} & \textbf{12.55\%} & \textbf{52.09\%} 
& \textbf{41.33\%} & \textbf{44.33\%} & \underline{34.67\%} 
& \textbf{8.02\%} & \textbf{12.66\%} & \textbf{74.26\%} 
& \textbf{43.80\%} & \textbf{61.40\%} \\
\bottomrule
\end{tabular}
}
\caption{Evaluation of \textbf{Joint Accuracy} across five datasets. Within each model group, \textbf{bold} denotes the best performance and ~\underline{underline} denotes the second-best.}
\label{tab:joint_acc_all}
\end{table*}

%% file: tables/evidence_score.tex
\begin{table}[!t]
\centering
\resizebox{1.0\linewidth}{!}{
\begin{tabular}{l c c c c c}
\toprule
\textbf{Model} & \textbf{FEVEROUS} & \textbf{EX-FEVER} & \textbf{FEVER} & \textbf{SciFACT} & \textbf{HOVER} \\
\midrule
GPT-4o & 0.4558 & 0.4185 & 0.4242 & 0.3187 & 0.6945 \\
\midrule
Qwen2.5-3B-Instruct & 0.2735 & 0.2893 & 0.2327 & 0.1940 & 0.5022 \\
Qwen2.5-3B-Instruct-SFT & 0.2468 & 0.2960 & 0.2397 & 0.1972 & 0.4753 \\
Qwen2.5-3B-Instruct-OfflineRL & \underline{0.3837} & \underline{0.3361} & \underline{0.3441} & \underline{0.2459} & 0.5389 \\
Qwen2.5-7B-Instruct & 0.3397 & 0.3323 & 0.3383 & 0.1891 & \underline{0.5646} \\
Qwen2.5-3B-Instruct-OnlineRL & \textbf{0.4769} & \textbf{0.4635} & \textbf{0.4630} & \textbf{0.2713} & \textbf{0.6562} \\
\midrule
Llama3.2-3B-Instruct & 0.1934 & 0.2444 & 0.1732 & 0.0963 & 0.3484 \\
Llama3.2-3B-Instruct-SFT & 0.1995 & 0.2433 & 0.1742 & 0.1004 & 0.3766 \\
Llama3.2-3B-Instruct-OfflineRL & 0.3097 & 0.3140 & 0.2601 & 0.0863 & 0.4587 \\
Llama3.1-8B-Instruct & \underline{0.3763} & \underline{0.3969} & \underline{0.3368} & \underline{0.2073} & \underline{0.6209} \\
Llama3.2-3B-Instruct-OnlineRL & \textbf{0.4607} & \textbf{0.4717} & \textbf{0.4389} & \textbf{0.2410} & \textbf{0.6610} \\
\bottomrule
\end{tabular}
}
\caption{Evidence Score Comparison across five datasets.
Within each model group, \textbf{bold} denotes the best performance and ~\underline{underline} denotes the second-best.}
\label{tab:evidence_scores}
\vspace{-1em}
\end{table}

%% file: sections/5_analysis.tex
\section{Analysis}
\subsection{Effect of Evidence Reward}
To directly incentivize precise evidence selection, we introduce an evidence‐score based reward during the online reinforcement‐learning phase. Figure~\ref{fig:evid-qwen} and~\ref{fig:evid-llama} compare the evolution of this metric with and without the evidence reward on the Qwen2.5‑3B‑Instruct model. While the “no‐reward” variant exhibits an initial uptick in evidence score over the first 40 training steps, its performance subsequently degrades. By contrast, the model trained with the evidence‐score reward maintains consistent improvement, demonstrating that explicit supervision via the evidence score is essential for robust, high‐precision evidence retrieval in an interactive verification setting.

\input{tables/ablation_study}
\subsection{Effect of Weight Validity}
To encourage LLMs to learn from genuinely correct reasoning rather than shortcut paths that merely happen to yield the right answer, we introduce a \textbf{validity weight}. We evaluate two key metrics during training: \textbf{evidence cover rate} and \textbf{verification accuracy}. The evidence cover rate is defined as the proportion of responses in which the model covers all gold-standard evidences.

As illustrated in Figure~\ref{fig:veri_acc-qwen} and~\ref{fig:veri_acc-llama}, both the Qwen and Llama models augmented with validity weight achieve higher verification accuracy throughout training. Because we incorporate an evidence-based reward into the total objective, models receive greater reward when they retrieve more gold evidences. As a result, they learn to produce answers grounded in the gold evidence set; yet, even with this evidence reward, they still underperform the variants trained with validity weight.

We also observe notable differences in evidence cover rate in Figure~\ref{fig:evid_cover-qwen} and~\ref{fig:evid_cover-llama}. For the Qwen model without validity weight, the evidence cover rate declines sharply after sixty training steps, implying that the model has discovered a shortcut to verify claims rather than performing the comprehensive verification we intended. Although this model gradually improves its ability to retrieve gold evidences, it remains inferior to the validity-weighted variant.

In conclusion, incorporating a validity weight guides the model toward reasoning trajectories that yield both correct answers and sound justifications, thereby stabilizing the training process and preventing the model from exploiting the reward function.

\input{tables/logit_analysis}
\subsection{Relationship between Accuracy and Confidence}
We conducted an experiment to quantify model confidence by extracting the output logit corresponding to each predicted label and recording these as triplets of label–logit pairs. Examination of the resulting logit distribution (Figure~\ref{fig:dist-qwen} and~\ref{fig:dist-llama}) reveals that most values cluster near zero, suggesting that, in general, models exhibit high confidence in their predictions. We attribute this phenomenon, at least in part, to the fact that extensive intermediate reasoning processes help models resolve uncertainty before producing a final judgment.

To investigate the relationship between answer confidence and accuracy, we defined two confidence tiers: low confidence (logit < 0.85) and high confidence (logit > 0.95). As shown in Figure ~\ref{fig:acc-con-qwen} and~\ref{fig:acc-con-llama}, for SUPPORT and REFUTE labels, both Qwen and Llama models demonstrate higher confidence levels accompanied by correspondingly higher accuracy rates. In contrast, the NOT ENOUGH INFO (NEI) label exhibits lower logits overall, and notably, higher confidence does not consistently translate into improved accuracy for this category. Indeed, the trend reverses for Qwen and Llama in this case.

We further observe in Figure ~\ref{fig:precision} and~\ref{fig:recall} that increasing model scale correlates with a decreased propensity to predict the “NOT ENOUGH INFO” category: in both architectures, the 7 and 8 billion-parameter variants exhibit lower NEI recall than their 3 billion-parameter counterparts.  In particular, the notably poor NEI recall of Qwen2.5-7B-Instruct appears to result from a semantic confusion in which many genuine NEI instances are misclassified as REFUTE.

%% file: tables/ablation_study.tex
\begin{figure}[htbp]
  \centering
    \begin{subfigure}[b]{0.49\linewidth}
    \centering
    \includegraphics[width=\linewidth]{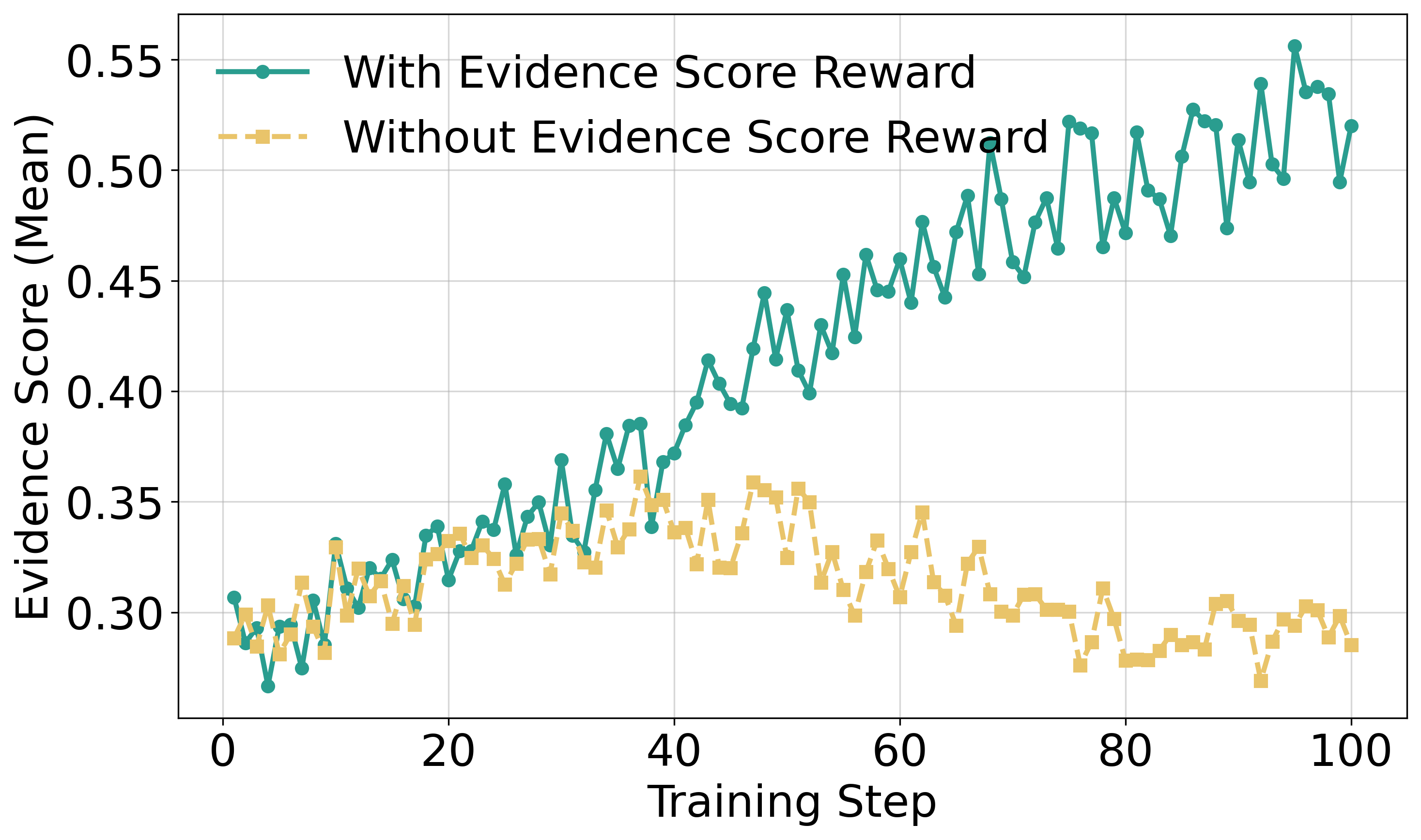}
    \caption{Qwen Evidence Score}
    \label{fig:evid-qwen}
  \end{subfigure}
  \hfill
  \begin{subfigure}[b]{0.49\linewidth}
    \centering
    \includegraphics[width=\linewidth]{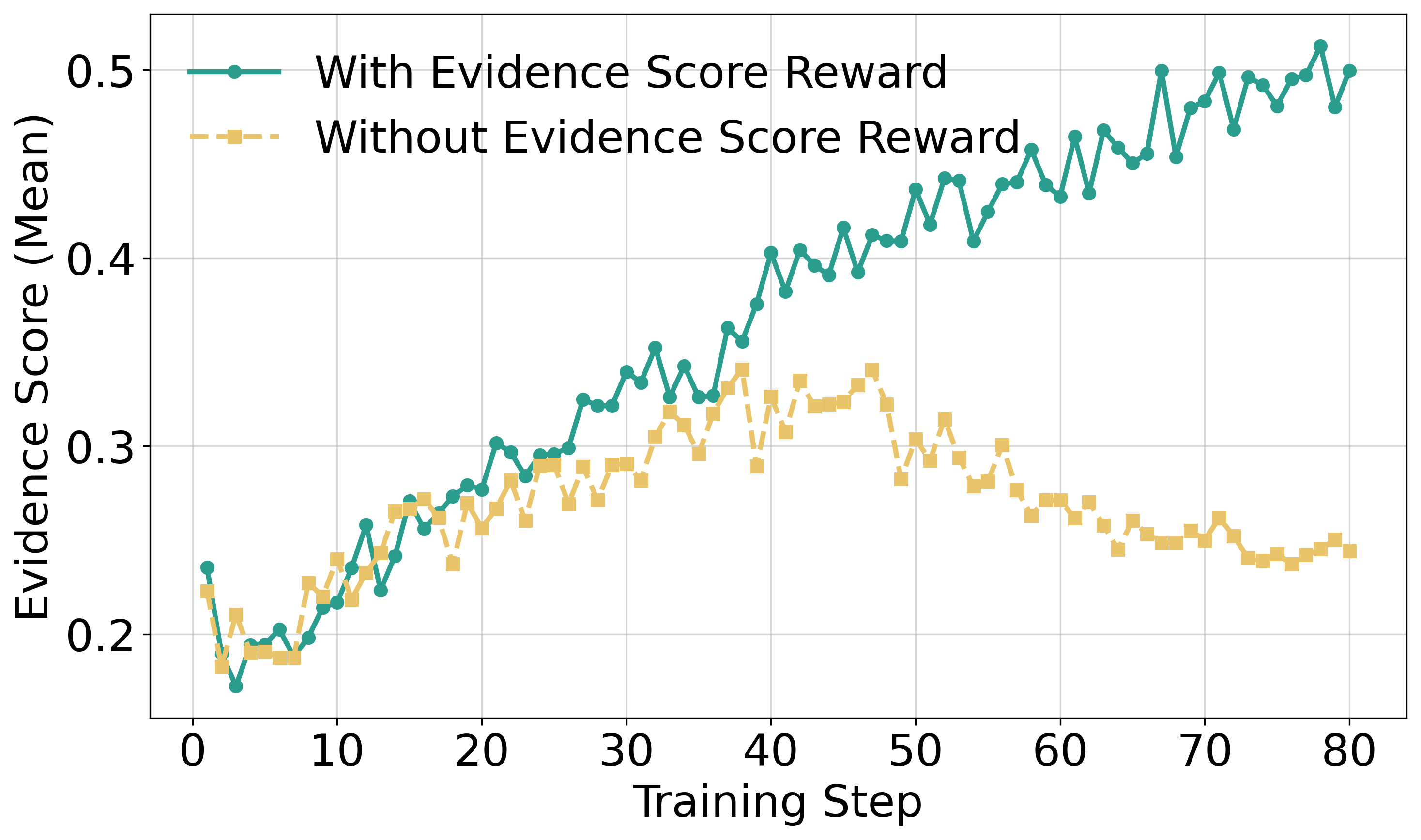}
    \caption{Llama Evidence Score}
    \label{fig:evid-llama}
  \end{subfigure}
  \begin{subfigure}[b]{0.49\linewidth}
    \centering
    \includegraphics[width=\linewidth]{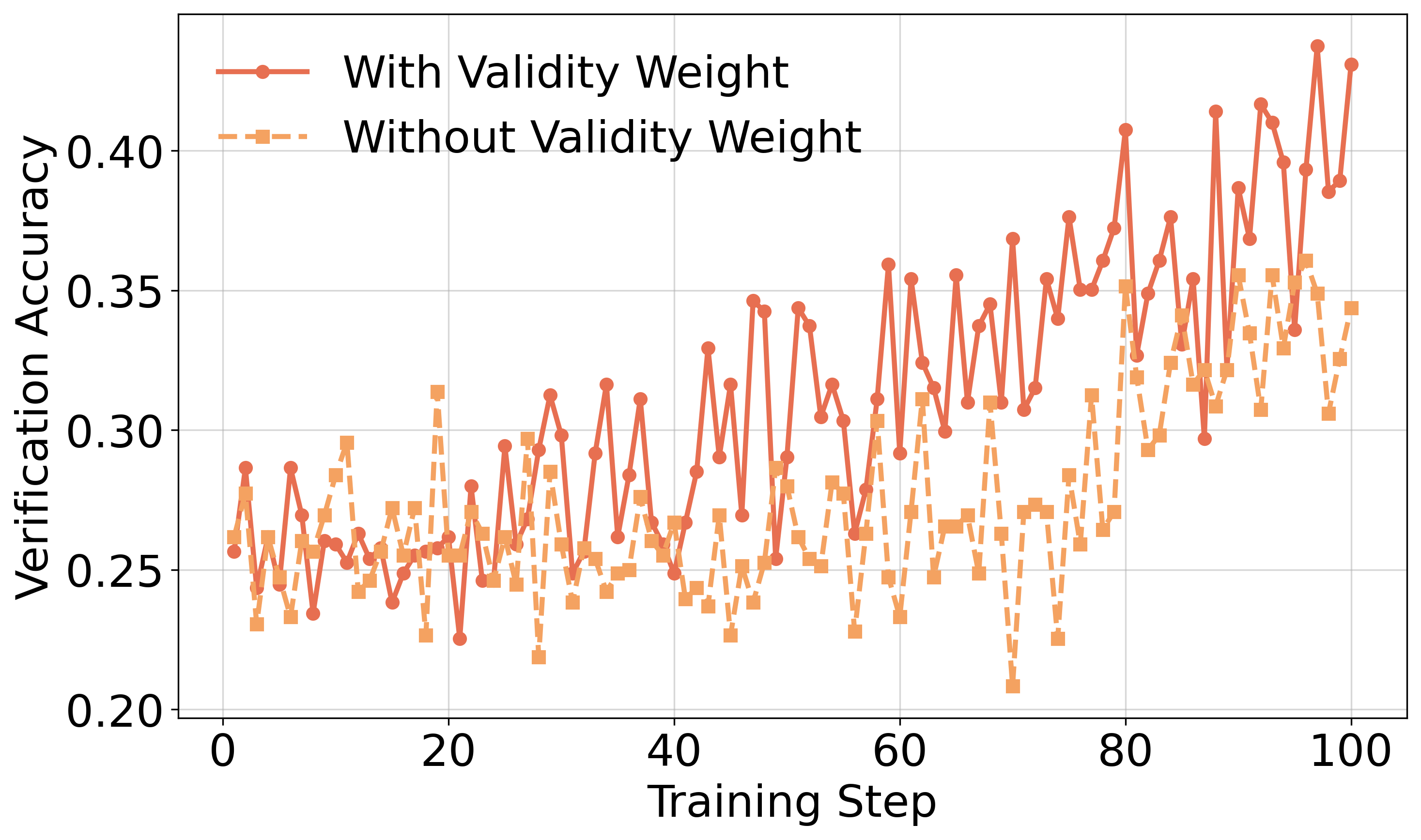}
    \caption{Qwen Veri Acc}
    \label{fig:veri_acc-qwen}
  \end{subfigure}
  \hfill
  \begin{subfigure}[b]{0.49\linewidth}
    \centering
    \includegraphics[width=\linewidth]{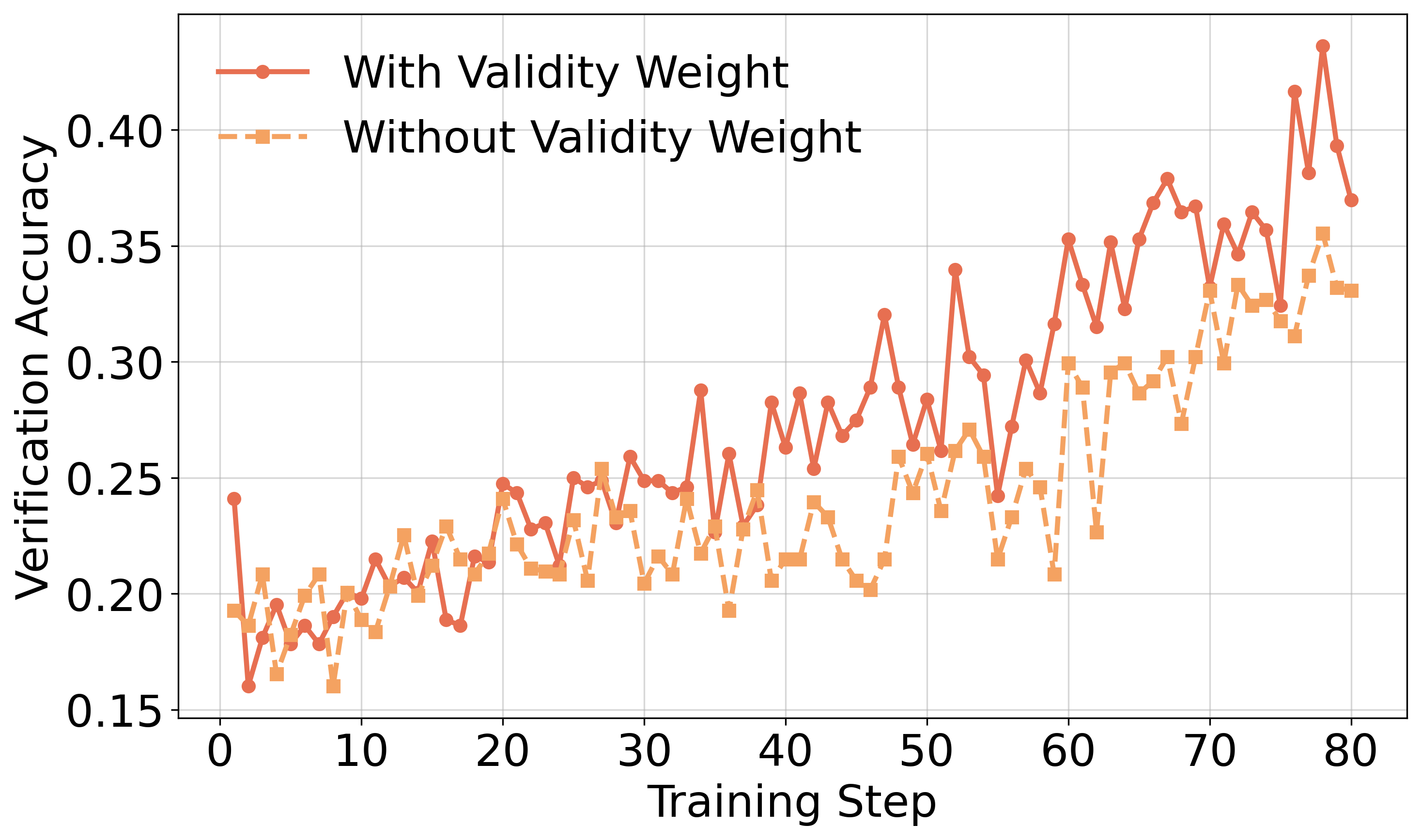}
    \caption{Llama Veri Acc}
    \label{fig:veri_acc-llama}
  \end{subfigure}
  \hfill
  \begin{subfigure}[b]{0.49\linewidth}
    \centering
    \includegraphics[width=\linewidth]{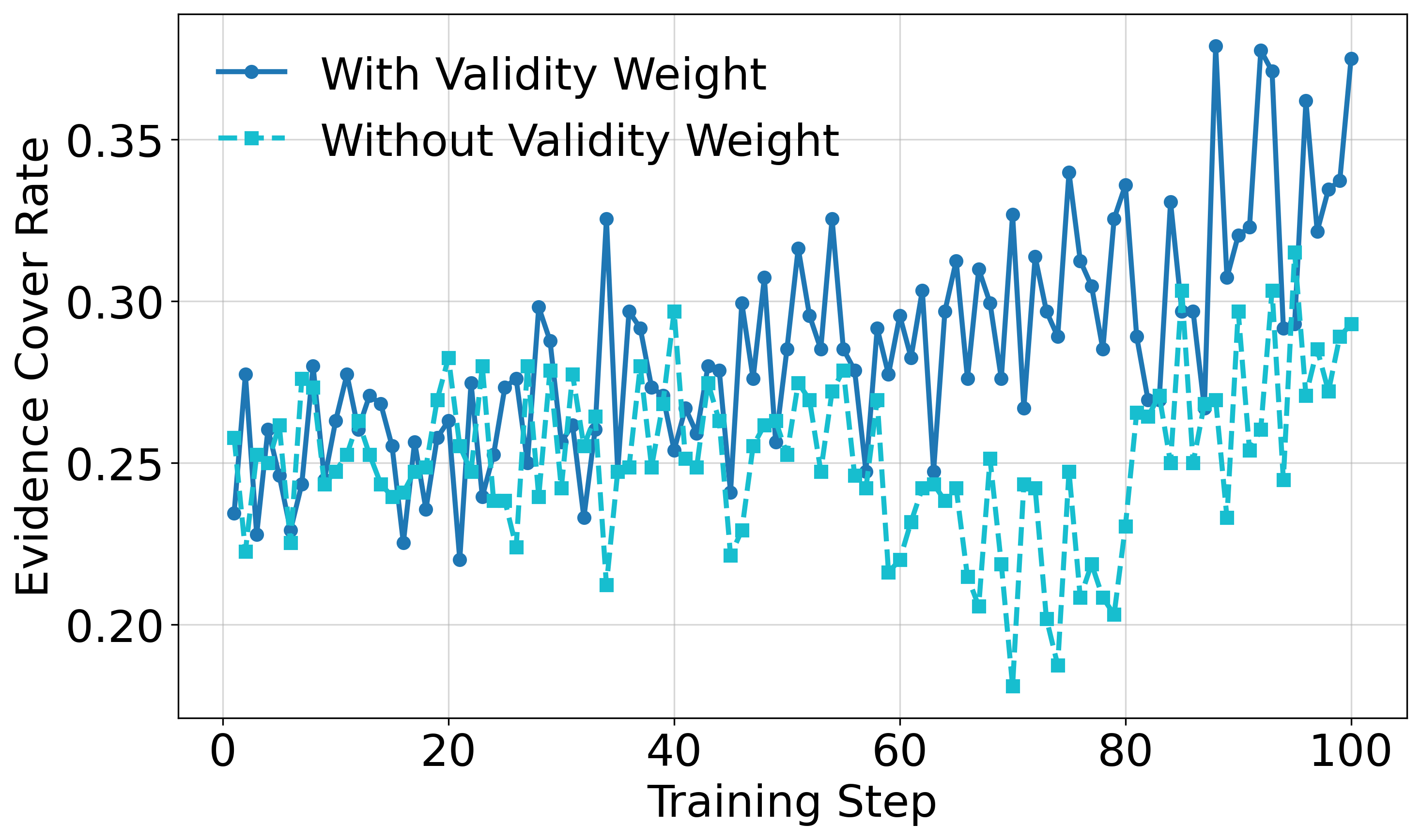}
    \caption{Qwen Evid Cover Rate}
    \label{fig:evid_cover-qwen}
  \end{subfigure}
  \begin{subfigure}[b]{0.49\linewidth}
    \centering
    \includegraphics[width=\linewidth]{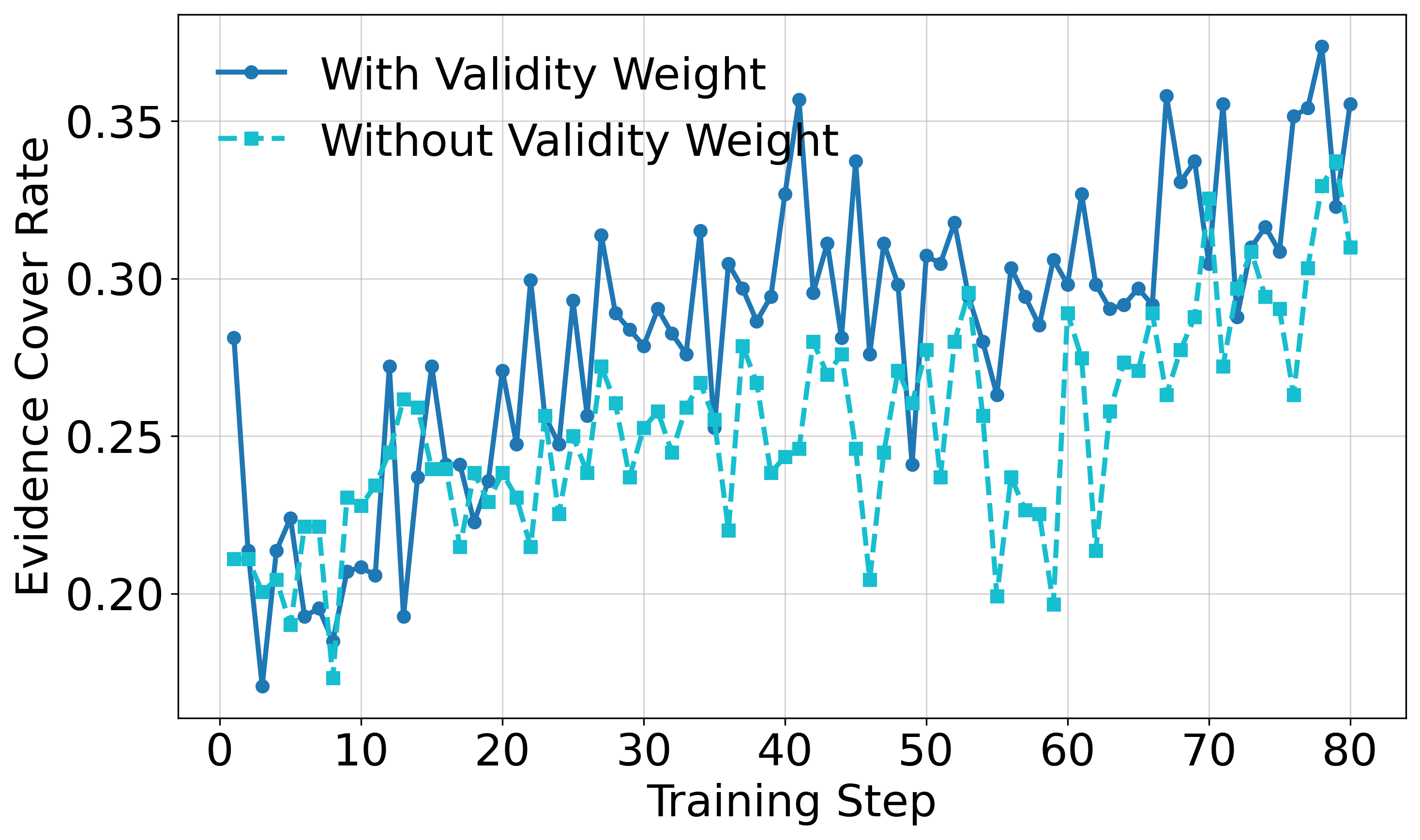}
    \caption{Llama Evid Cover Rate}
    \label{fig:evid_cover-llama}
  \end{subfigure}
  \caption{Training curves from the ablation study. Panels (a)–(b) report evidence score under the evidence score ablation, while panels (c)–(d) show verification accuracy and panels (e)–(f) show evidence cover rate under the validity weight ablation.}
  \label{fig:ablation_study}
\end{figure}

%% file: tables/logit_analysis.tex
\begin{figure}[htbp]
  \centering
    \begin{subfigure}[b]{0.48\linewidth}
    \centering
    \includegraphics[width=\linewidth]{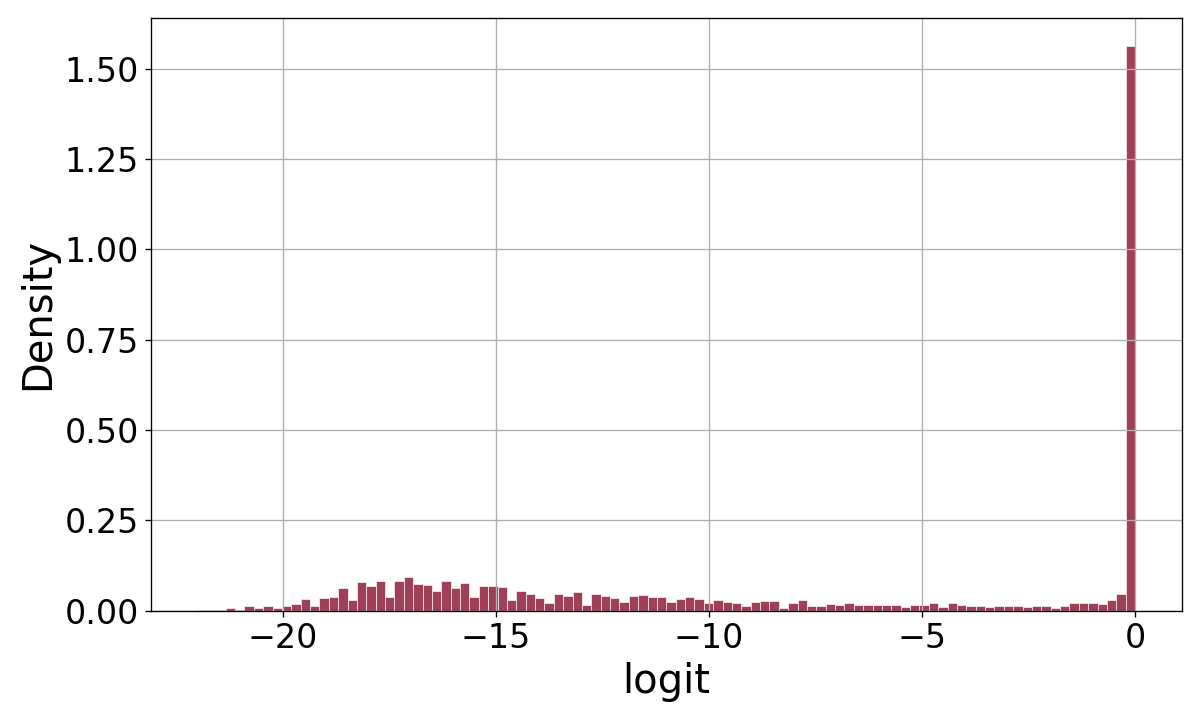}
    \caption{Qwen Logit Dist}
    \label{fig:dist-qwen}
  \end{subfigure}
  \hfill
  \begin{subfigure}[b]{0.48\linewidth}
    \centering
    \includegraphics[width=\linewidth]{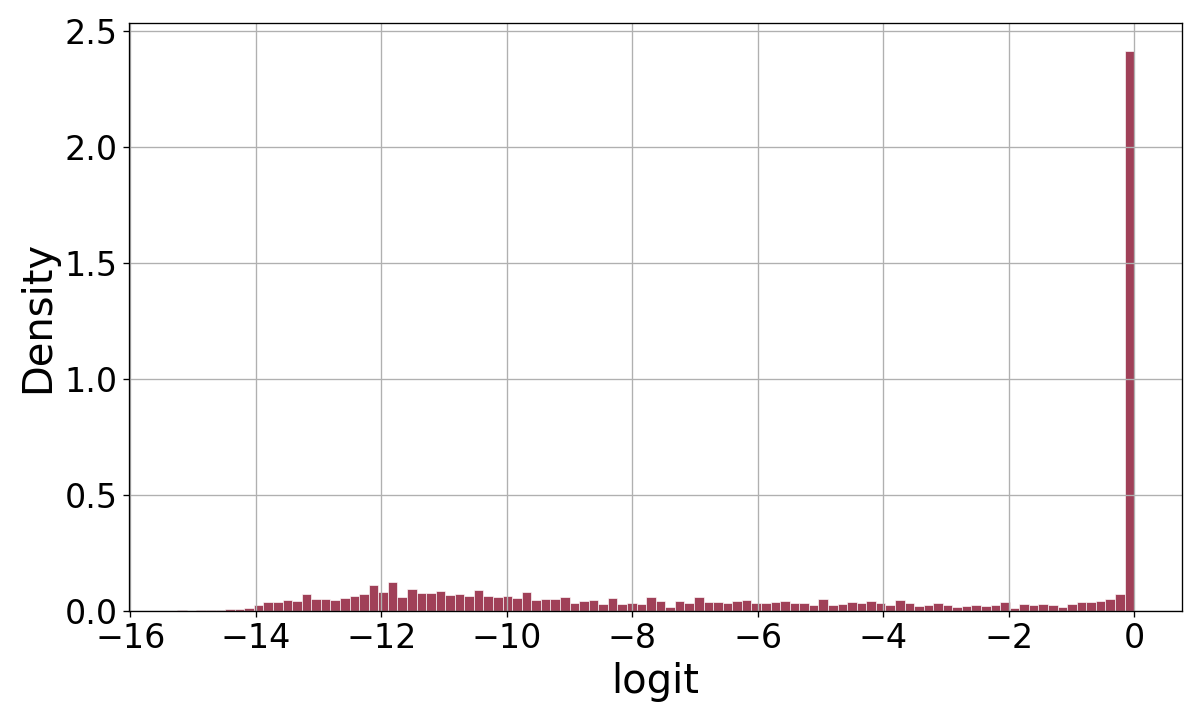}
    \caption{Llama Logit Dist}
    \label{fig:dist-llama}
  \end{subfigure}
  \begin{subfigure}[b]{0.48\linewidth}
    \centering
    \includegraphics[width=\linewidth]{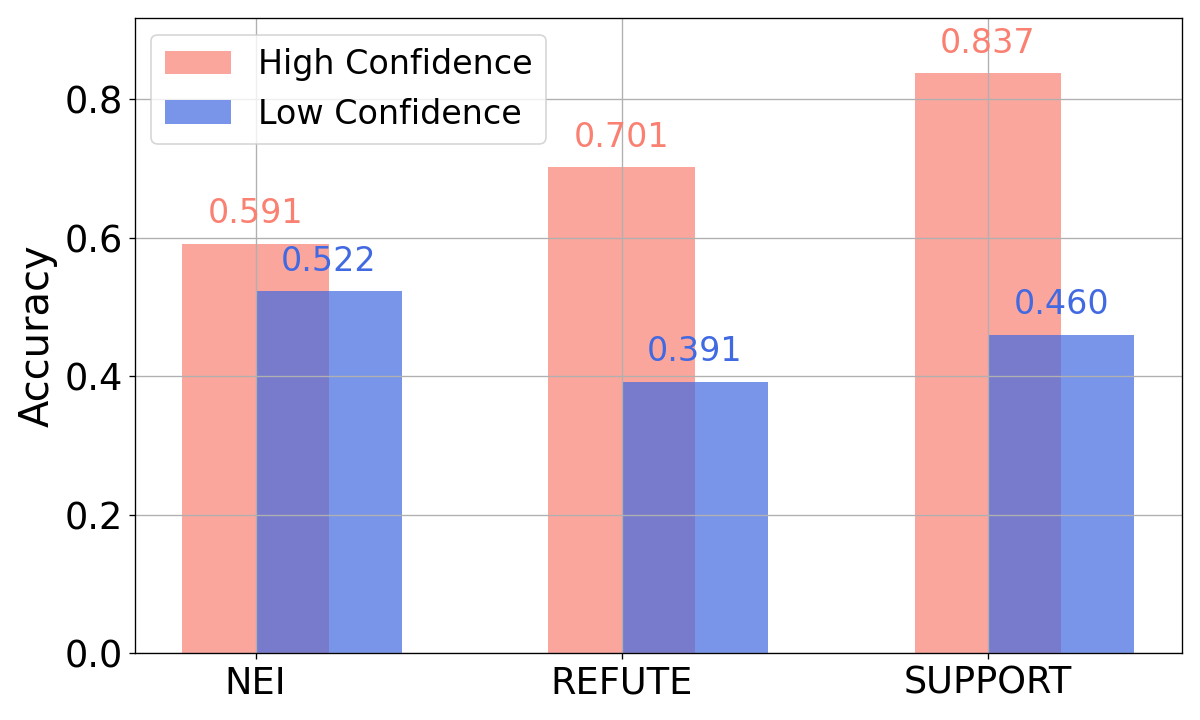}
    \caption{Qwen Acc vs.\ Conf}
    \label{fig:acc-con-qwen}
  \end{subfigure}
  \hfill
  \begin{subfigure}[b]{0.48\linewidth}
    \centering
    \includegraphics[width=\linewidth]{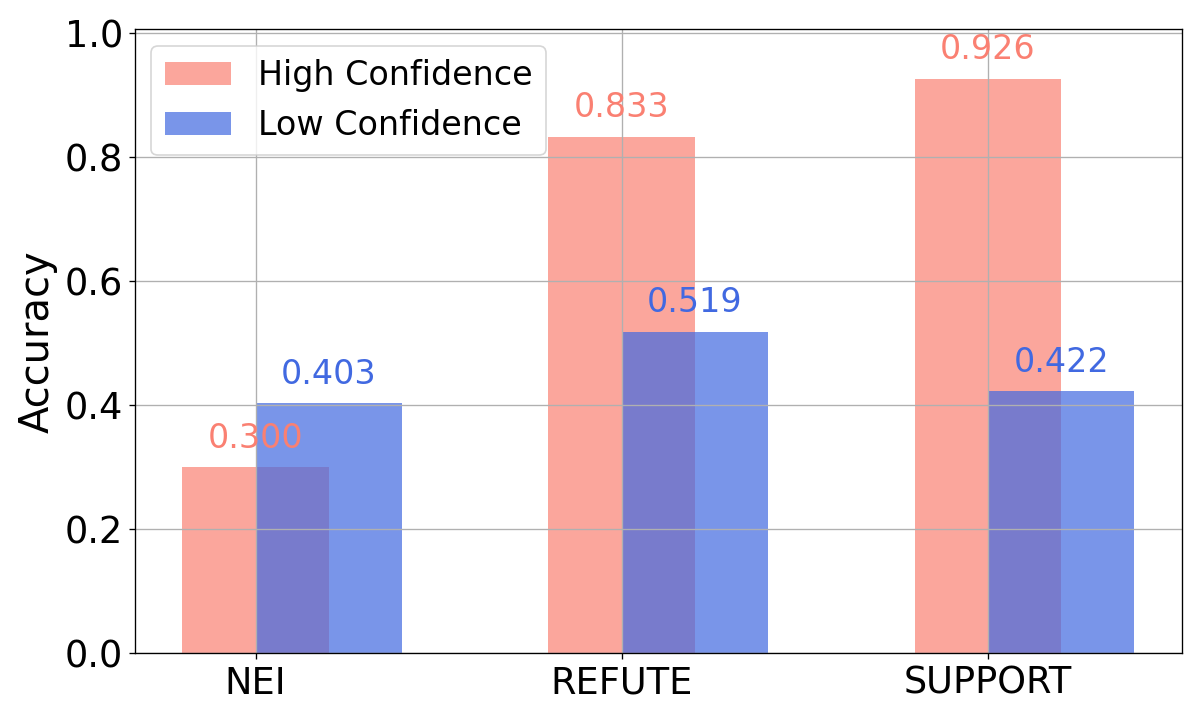}
    \caption{Llama Acc vs.\ Conf}
    \label{fig:acc-con-llama}
  \end{subfigure}
  \hfill
  \begin{subfigure}[b]{0.48\linewidth}
    \centering
    \includegraphics[width=\linewidth]{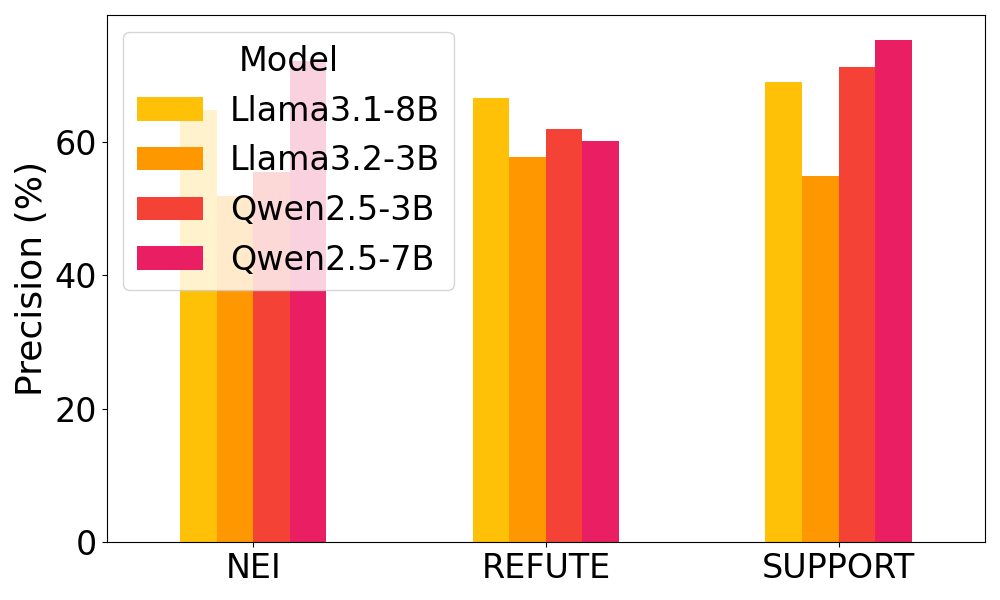}
    \caption{Precision Analysis}
    \label{fig:precision}
  \end{subfigure}
  \begin{subfigure}[b]{0.48\linewidth}
    \centering
    \includegraphics[width=\linewidth]{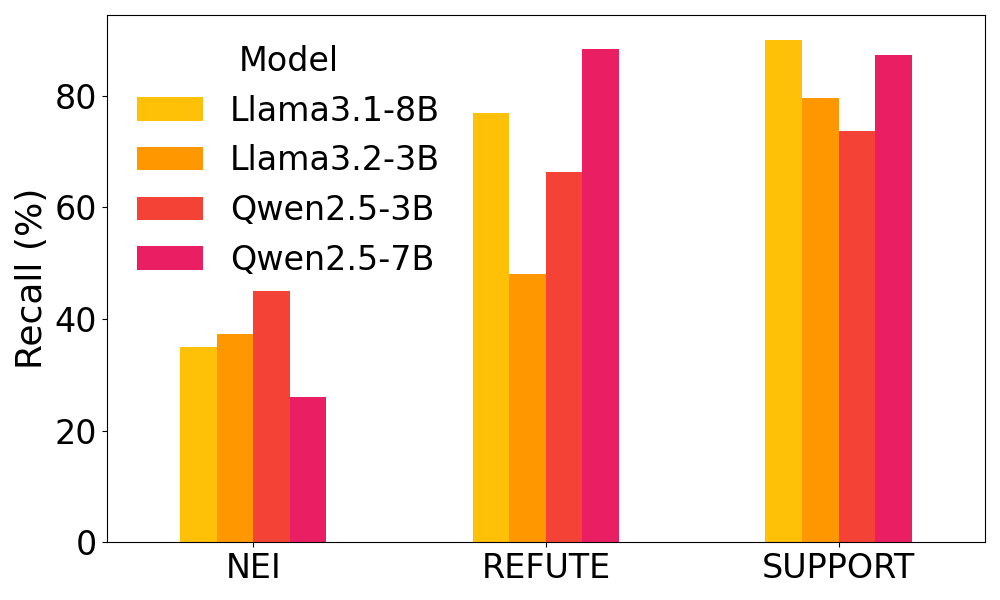}
    \caption{Recall Analysis}
    \label{fig:recall}
  \end{subfigure}
  \caption{Bar-based analysis of accuracy and logits. Panels (a)–(b) display the logit distributions of Qwen and Llama, panels (c)–(d) report accuracy versus confidence, and panels (e)–(f) present recall and precision across models and labels.}
  \label{fig:Logit_Analysis}
\end{figure}

%% file: sections/6_conclusion.tex
\section{Conclusion}
To address the critical challenge of enhancing verification capabilities across diverse scenarios, we introduce the \textbf{Veri-R1} framework, which leverages reinforcement learning to guide models in reasoning, evidence retrieval, and judgment under an online claim verification setting. Empirically, models trained with Online RL consistently outperform their counterparts of the same scale trained via SFT or Offline RL, and in many cases even surpass larger-scale models within the same series—demonstrating the effectiveness of the Veri-R1 paradigm. Our component-wise reward analysis elucidates the specific contributions of each reward signal to the training process, while logit probing reveals the relationship between output confidence and answer accuracy. We envision this work as a step toward more precise and faithful claim verification pipelines capable of handling diverse real-world challenges.
\subsection*{Limitations}
Our training and evaluation are performed in an online setting with a fixed local corpus and retriever. In practice, claim verification often operates over far larger, dynamic corpora with continuously emerging information. Integrating a real-world–scale retriever into this framework would better reflect practical conditions and likely improve the reliability and robustness of both training and evaluation.

%% file: sections/7_appendix.tex
\cleardoublepage
\appendix

\section{Logit Distribution for Each Label}

For the \textbf{SUPPORT} label, the answer logits produced by both models are primarily concentrated near 1, indicating high model confidence. Additionally, there is a secondary, low-density concentration in the lower logit regions — approximately between -20 and -10 for Qwen, and between -14 and -10 for LLaMA. As previously analyzed, answers falling within these low logit intervals are more likely to be incorrect.
\begin{figure}[htbp!]
    \centering
    \includegraphics[width=0.49\linewidth]{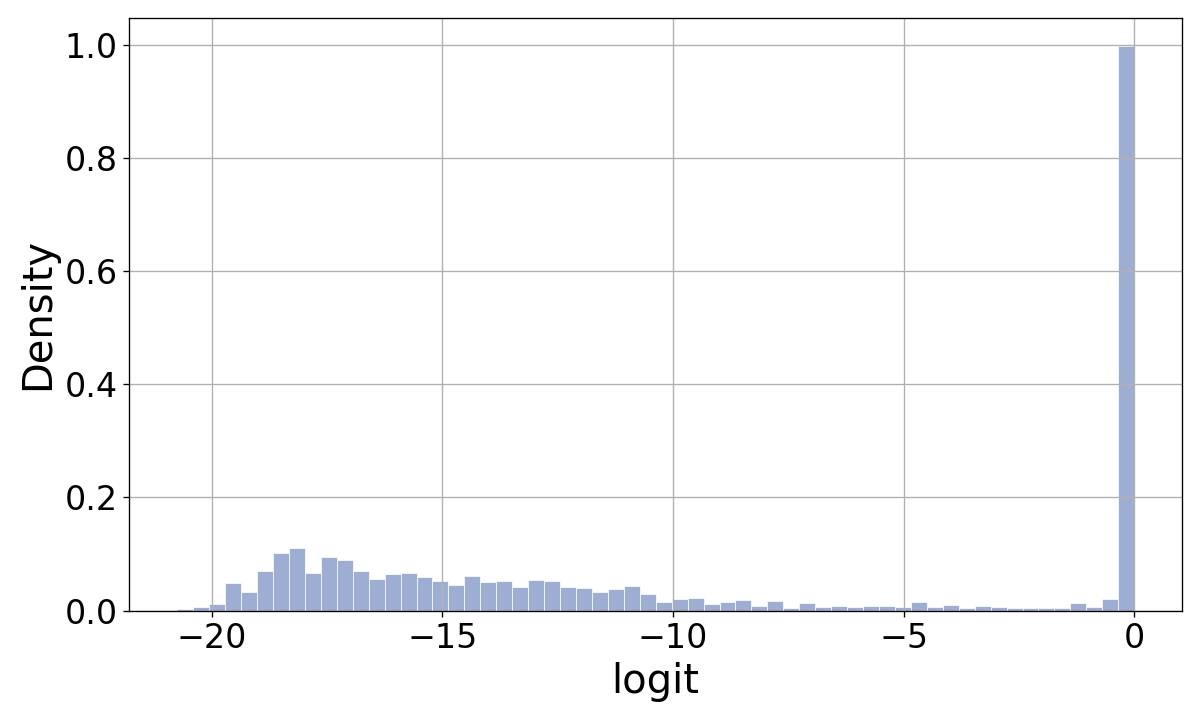}
    \includegraphics[width=0.49\linewidth]{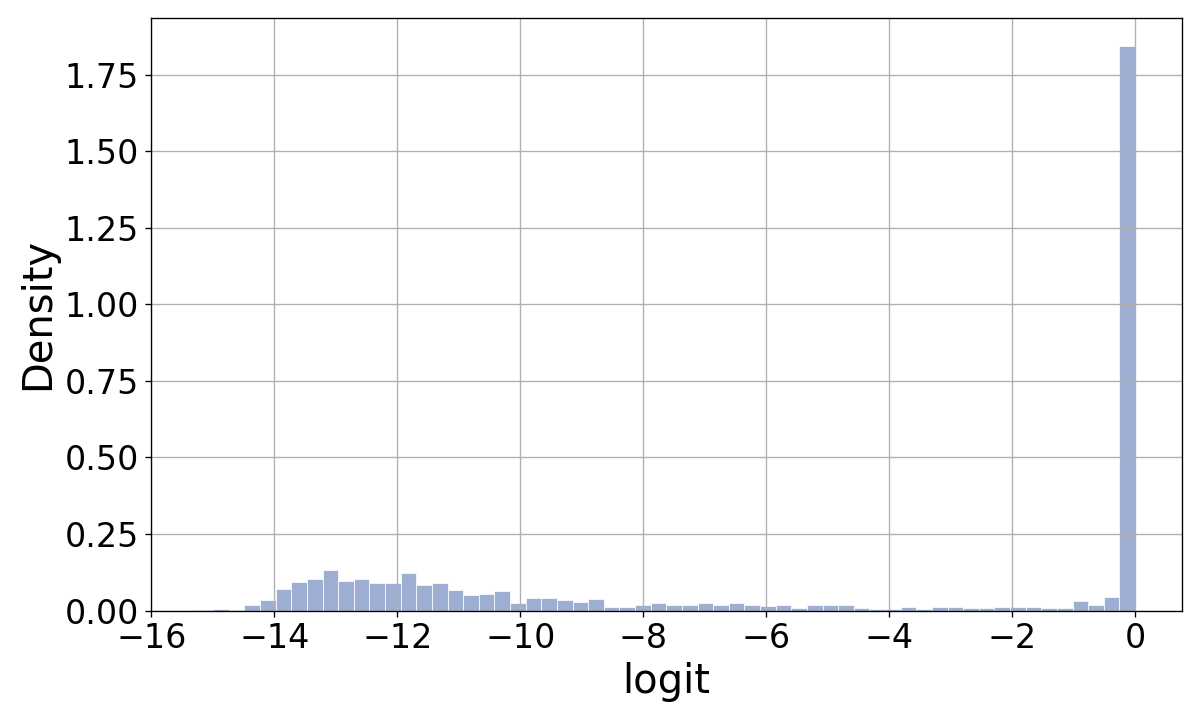}
    \caption{Logit Distribution for SUPPORT}
    \label{fig:label_logit_support}
\end{figure}

For the \textbf{REFUTE} label, the logits are similarly centered near 1, reflecting high confidence for correct predictions. However, unlike the SUPPORT label, there is no notable secondary peak; instead, the remaining logits are relatively evenly distributed across the entire range, suggesting a more uniform uncertainty profile.
\begin{figure}[htbp!]
    \centering
    \includegraphics[width=0.49\linewidth]{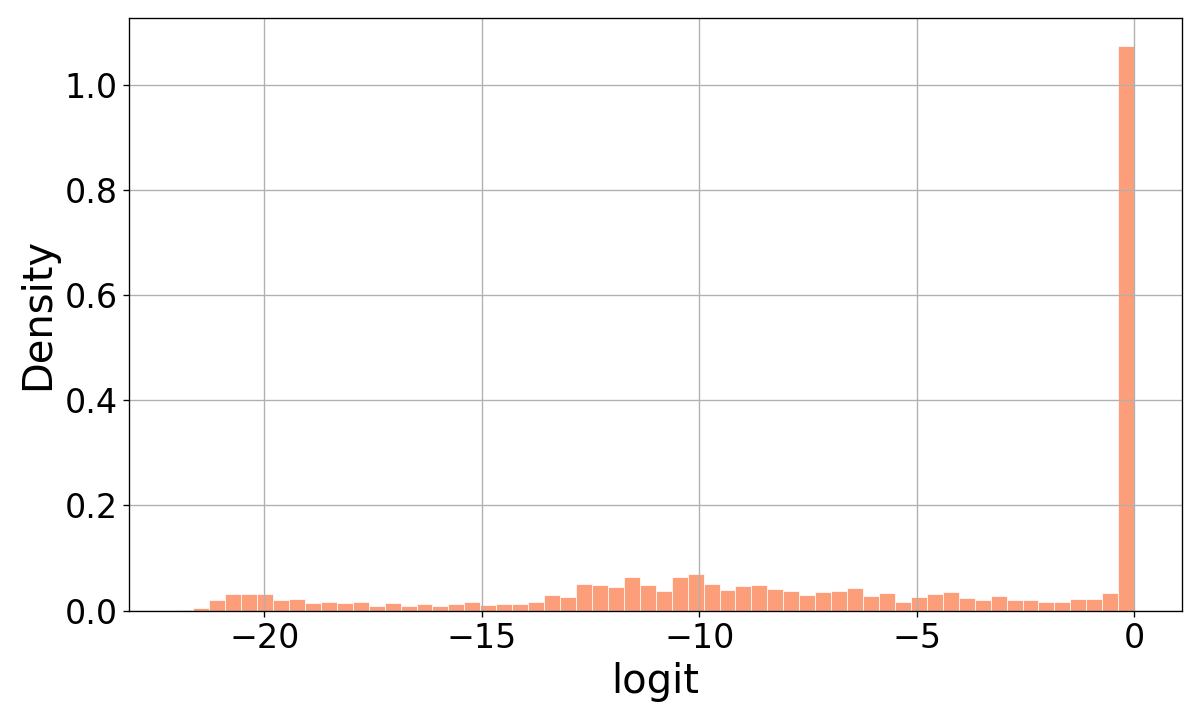}
    \includegraphics[width=0.49\linewidth]{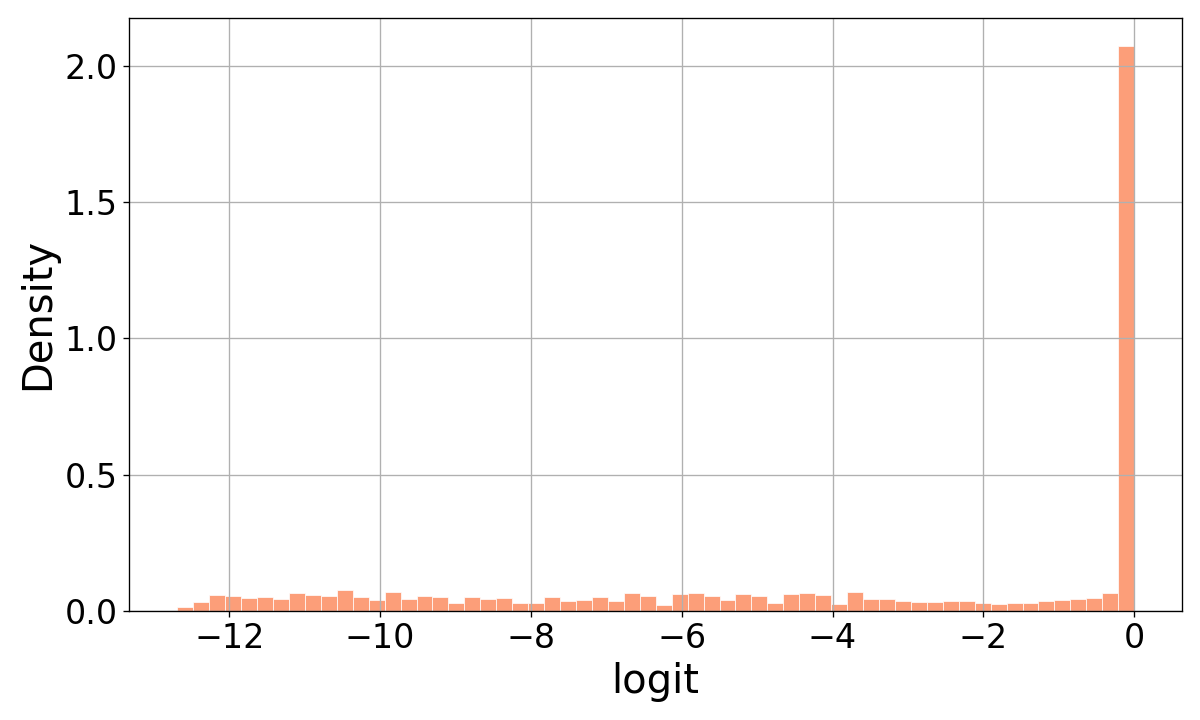}
    \caption{Logit Distribution for REFUTE}
    \label{fig:label_logit_refute}
\end{figure}

In contrast, for the \textbf{NOT ENOUGH INFO} label, the models exhibit substantially lower confidence. This is evident from the noticeably reduced density near logit = 1. Furthermore, the logits are more widely spread across the range, with elevated densities in non-peak regions compared to the SUPPORT and REFUTE labels. This indicates a broader distribution of model uncertainty when handling instances labeled as NOT ENOUGH INFO.
\begin{figure}[htbp!]
    \centering
    \includegraphics[width=0.49\linewidth]{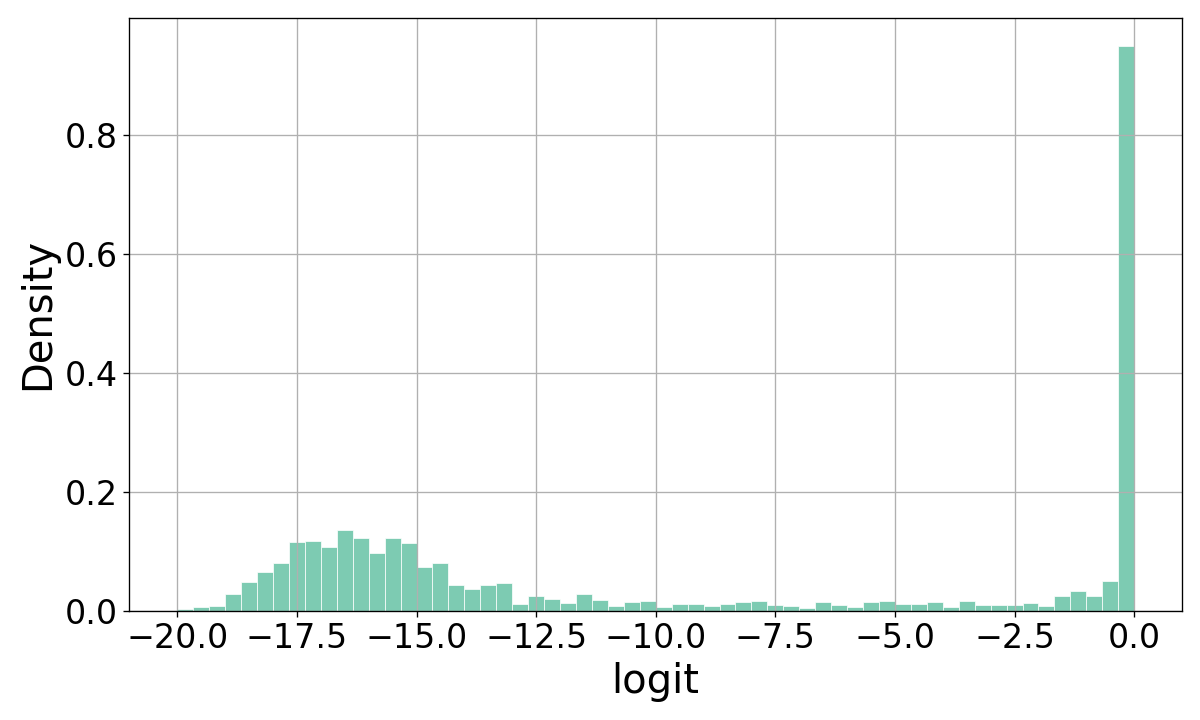}
    \includegraphics[width=0.49\linewidth]{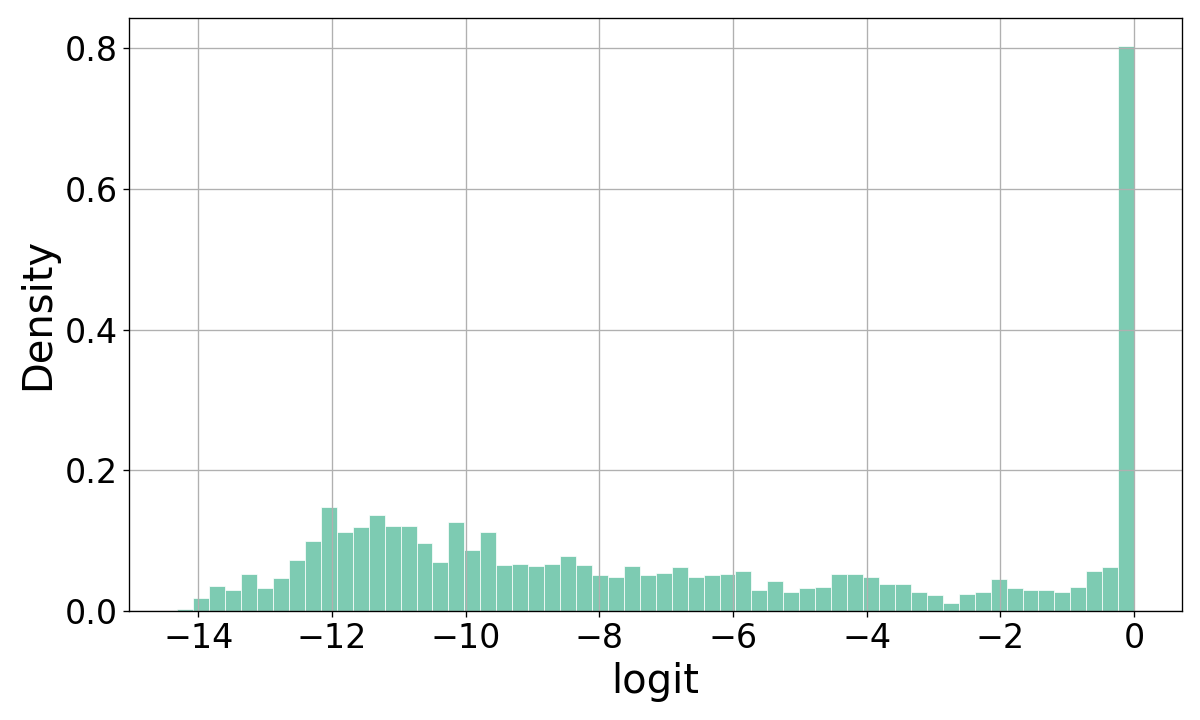}
    \caption{Logit Distribution for NEI}
    \label{fig:label_logit_nei}
\end{figure}

\section{Training Configuration}
All experiments were conducted on a compute node with two NVIDIA A800-80G GPUs. Running Online RL training on this hardware takes approximately 15 hours to complete 100 steps. Sentence retrieval from the corpus is performed using the FAISS library. In addition, we group every three sentences into one entry, and in each turn we retrieve the most related three entries.
\input{tables/training_setting}

\section{Details of Dataset Processing}\label{appendix:dataset}
\paragraph{FEVEROUS} To counteract the imbalance in both task difficulty and label distribution, we include every sample from underrepresented challenge categories and uniformly subsample 3,000 instances for each label before adding them to our training pool. Besides, we evenly sampled data each label from the development set as evaluation set.
\paragraph{EX‑FEVER} In our experiments, we utilize only the gold labels and evidence, randomly selecting 3,000 examples per label for inclusion in the training pool. Besides, we evenly sampled data each label from the development set as evaluation set. All the data of EX-FEVER comes from the development set.

\paragraph{Final Training Dataset}From the aggregated pool of FEVEROUS and EX-FEVER, we first apply our filtering pipeline in Figure~\ref{fig:data_filter} to remove low‑quality instances, then resample from the filtered data to assemble the final training set.

\paragraph{FEVER} We randomly select 900 data samples from the original FEVER dataset, comprising 300 instances for each label. As the gold-standard evidence in FEVER is annotated at the sentence level, we preprocess the source documents by splitting them into individual sentences and grouping every three consecutive sentences to construct compact retrieval units.

\paragraph{SciFACT} Due to the limited size of the SciFACT dataset, we select as many samples as available for the label with the fewest instances (237 samples each label). To ensure label consistency across datasets, we map the label CONTRADICT to REFUTE, aligning it with the labeling scheme used in FEVER. The document preprocessing procedure follows the same approach as in FEVER.

\paragraph{HOVER} As a multi-hop fact verification dataset, HOVER contains a considerable proportion of ambiguous claims. To address this, we employ GPT-4o to filter and retain samples with clearer, more objective judgments. Additionally, given that multi-hop claims often require extensive evidence, which poses challenges for models in accurately retrieving all supporting sentences, we limit our evaluation to 2-hop claims. The evaluation set is curated to ensure a balanced distribution between the SUPPORT and NOT SUPPORTED labels (448 samples each label).

\section{Data Quality}
Data quality plays a crucial role in maintaining a stable training process and ensuring the reliability of evaluation outcomes. During our data preparation stage, we identified several types of issues and ambiguities within the raw datasets. To illustrate the potential limitations inherent in the current data, we selected representative examples of these problems. Highlighting such issues not only reveals challenges in the existing datasets but also provides guidance for improving data creation practices in future work.
\input{tables/data_quality}

\section{Prompt}
The system prompt for online claim verification and offline claim verification is respectively illustrated in Figure~\ref{tab:Online_RL} and~\ref{tab:Offline_RL}. Similar to the instruction in Figure~\ref{tab:Online_RL} for online setting, prompt in Figure~\ref{tab:Offline_RL} is designed to leverage reinforcement learning (RL) for training LLMs in an offline setting. In this framework, the models are required solely to perform reasoning and provide answers.

\input{tables/offline_prompt}

\section{Additional Results}
To comprehensively evaluate models, we also report label-wise verification accuracy in Figures~\ref{tab:veri_results_all}, and label-wise label accuracy in Figures~\ref{tab:label_results_all}. Online RL-trained models achieve the highest verification accuracy compared to other models of the same scale, and their performance is comparable to that of 7B and 8B counterparts. In terms of label accuracy, although Online RL does not exhibit overwhelmingly superior results, it consistently outperforms the baseline instruct models. Offline RL demonstrates particular effectiveness in improving label accuracy for the Qwen model. However, label accuracy may not always provide a fully reliable measure, as models can achieve high scores in specific labels while relying on incorrect reasoning, reflecting inherent biases or preferences.
\input{tables/veri_acc}
\input{tables/label_acc}

\section{Case Analysis}

\begin{figure*}[!t]
    \centering
    \includegraphics[width=0.85\linewidth]{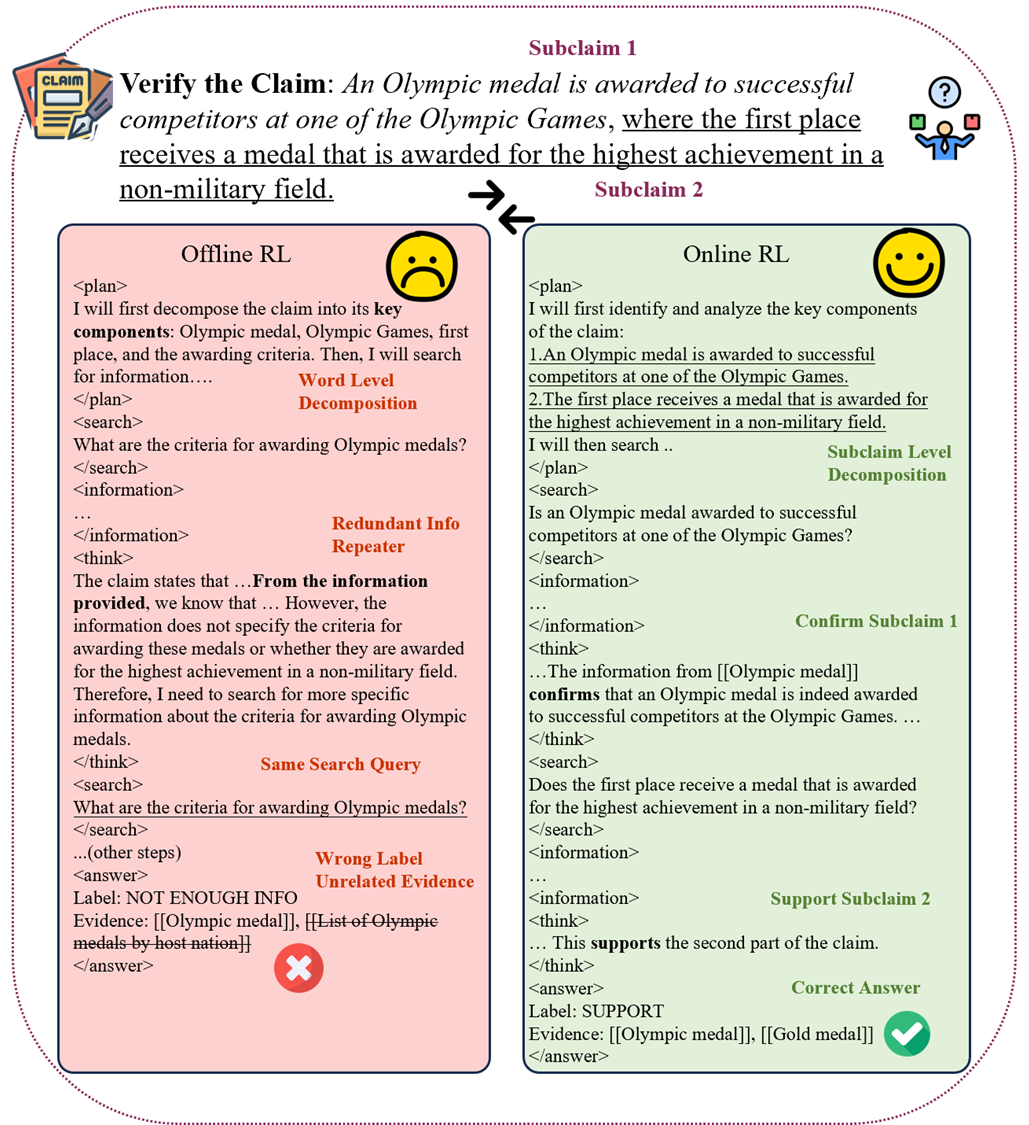}
    \caption{Case Analysis of Offline RL V.S. Online RL}
    \label{fig:case_analysis}
\end{figure*}

In comparing the two RL trained models’ outputs (Figure~\ref{fig:case_analysis}), two key dimensions emerge: \textbf{decomposition granularity} and \textbf{retrieval–evidence utilization}.

\subsection*{Decomposition Granularity}

\paragraph{Offline RL - Coarse (word‑level) decomposition} 
The first model extracts isolated keywords — such as “Olympic medal”, “Olympic Games”, “first place”, and “awarding criteria” — and issues a single undifferentiated query. This strategy overlooks the claim’s internal logical structure and impedes precise verification of individual subclaims.

\paragraph{Online RL - Fine‑grained (subclaim‑level) decomposition} 
By contrast, the second model partitions the overall assertion into two subclaims:
\begin{enumerate}
  \item “An Olympic medal is awarded to successful competitors at one of the Olympic Games.”
  \item “First place receives a medal awarded for the highest achievement in a non‑military field.”
\end{enumerate}
Separate and tailored searches are conducted for each subclaim, ensuring that each component of the original claim undergoes independent validation.

\subsection*{Retrieval and Evidence Utilization}

\paragraph{Offline RL - Redundant retrieval and unfocused evidence} 
The first model repeatedly issues the same generic query (e.g., What are the criteria for awarding Olympic medals ?), obtains relevant or background information, and ultimately produces a “\textsc{Not Enough Info}” label with irrelevant evidence.

\paragraph{Online RL - Targeted retrieval and incremental confirmation} 
The second model formulates distinct queries for each subclaim and immediately integrates the retrieved information into its reasoning. After confirming Subclaim 1, it proceeds to Subclaim 2, thereby constructing a coherent chain of thought. This workflow culminates in the correct “\textsc{Support}” judgment supported by precisely relevant evidence (e.g., the Olympic medal awarding criteria and the definition of a gold medal).

Therefore, A subclaim‑based decomposition strategy—combined with targeted retrieval and stepwise evidence confirmation—substantially enhances the precision and reliability of fact‑verification performance.

%% file: tables/training_setting.tex
\begin{table}[ht]
\centering
\label{tab:training-params}
\begin{tabular}{ll}
\toprule
\multicolumn{2}{l}{\textbf{Setting}} \\ 
\midrule
Train Batch Size        & 256                                           \\
Validation Batch Size   & 256                                           \\
Max Prompt Length       & 4864                                          \\
Max Response Length     & 512                                           \\
Max Start Length        & 512                                           \\
Max Observation Length  & 768                                           \\
\midrule
\multicolumn{2}{l}{\textbf{Actor Rollout}} \\ 
\midrule
Learning Rate           & 1e-6                                          \\
LR Warmup Ratio         & 0.285                                         \\
Use KL Loss             & true                                          \\
PPO Mini Batch Size     & 64                                            \\
PPO Micro Batch Size    & 16                                            \\
KL Loss Coefficient     & 0.001                                         \\
KL Loss Type            & low\_var\_kl                                  \\
\midrule
\multicolumn{2}{l}{\textbf{Actor Rollout Reference — Rollout}} \\ 
\midrule
Log-Prob Micro Batch Size   & 64                                         \\
Tensor Model Parallel Size  & 1                                          \\
Rollout Engine Name         & vllm                                       \\
GPU Memory Utilization      & 0.6                                        \\
Number of Agents            & 3                                          \\
Temperature                 & 0.8                                        \\
\midrule
\multicolumn{2}{l}{\textbf{Actor Rollout Reference — Reference Model}} \\ 
\midrule
Ref Log-Prob Micro Batch Size & 64                                       \\
Ref FSDP Param Offload         & True                                     \\
\midrule
\multicolumn{2}{l}{\textbf{Trainer Setting}} \\ 
\midrule
GPUs per Node               & 2                                        \\
Save Frequency (epochs)     & 5                                        \\
Test Frequency (epochs)     & 5                                        \\
Total Epochs                & 10                                       \\
\midrule
\multicolumn{2}{l}{\textbf{Other}} \\ 
\midrule
Max Turns                   & 4                                        \\
Retriever Top-K             & 3                                        \\
\bottomrule
\end{tabular}
\caption{Configuration For Training Process}
\end{table}

%% file: tables/data_quality.tex
\newcommand{\green}[1]{\textcolor{deepgreen}{#1}}
\begin{table}[htbp]
\centering
\small
\renewcommand{\arraystretch}{1.0} 
\setlength{\tabcolsep}{4pt}       
\label{tab:data_quality_issues}
\begin{minipage}{0.49\textwidth}
\centering
\begin{tabularx}{\linewidth}{l X}
\hline
\textbf{Claim 1} & It is illegal in Illinois to record a conversation.\\
\textbf{Evidence 1} & [[Illinois law]]: Illinois law prohibits recording private conversations without the consent of all parties. \\
\textbf{Evidence 2} & [[2014 Revision]]: In 2014, the Illinois Supreme Court struck down the old statute, and the revised law now bans only secret recordings of private conversations. \\
\textbf{Label} & \textcolor{red}{\sout{REFUTE}} \green{NOT ENOUGH INFO}\\
\textbf{Note} & \underline{\textbf{Overgeneralization:}} The claim is too absolute. The evidence shows that the law only makes it illegal under specific conditions (i.e. for private conversations, lacking consent). It is not correct to say all conversations in Illinois may not be recorded.\\
\hline
\textbf{Claim 2} & Honda Racing is an annual event organized by Honda to publicize ongoing projects.\\
\textbf{Evidence 1} & [[Honda Winner\_sentence\_0]]: The Honda Winner is an underbone motorcycle from the Japanese manufacturer Honda.\\
\textbf{Label} & \textcolor{red}{\sout{REFUTE}} \green{NOT ENOUGH INFO}\\
\textbf{Note} & \underline{\textbf{Entity mismatch:}} Evidence only describes Honda Winner as a motorcycle, not an event, leaving the claim unverified.\\
\hline
\textbf{Claim 3} & Jennifer Garner is an American actress raised in a city that is located at the confluence of the Elk and Kanawha rivers.\\
\textbf{Evidence 1} & [[Jennifer Garner]]: Jennifer Anne Garner (born April 17, 1972) is an American actress. Her breakthrough film debut was in the comedy \textit{Dude, Where's My Car} (2000). Following a supporting role in \textit{Pearl Harbor} (2001), Garner gained recognition for her performance as CIA officer Sydney Bristow in the ABC spy-action thriller \textit{Alias}, which aired from 2001 to 2006. For her work on the series, she won a Golden Globe Award and a SAG Award and received four Emmy Award nominations. \\
\textbf{Evidence 2} & [[Charleston, West Virginia]]: Charleston is the capital and the largest city in the U.S. state of West Virginia, and the county seat of Kanawha County. It is located at the confluence of the Elk and Kanawha Rivers in Kanawha County.\\
\textbf{Label} & \textcolor{red}{\sout{SUPPORT}} \green{NOT ENOUGH INFO}\\
\textbf{Note} & \underline{\textbf{Incomplete context:}} While the city’s location is confirmed, there is no direct evidence that Garner was raised there.\\
\hline
\end{tabularx}
\end{minipage}
\caption{Examples of Data Quality Issues in Claim Verification}
\end{table}
The table illustrates three common data quality issues in claim verification: (1) \textbf{Overgeneralization} where the claim makes an overly absolute statement not fully supported by the evidence;  (2) \textbf{Entity mismatch} where the subject in the claim is different from the evidence; and (3) \textbf{Incomplete context} where critical linking information is missing. Addressing these issues improves label accuracy and dataset reliability.

%% file: tables/offline_prompt.tex
\begin{figure*}[!t]
\centering
\resizebox{0.85\textwidth}{!}{%
\begin{tcolorbox}[colback=yellow!5!white, colframe=green!75!black,
  title=Claim Verification Assistant Prompt, boxrule=0.3mm,
  width=\textwidth, arc=3mm, auto outer arc=true]

You are a claim-verification assistant. You MUST follow this protocol exactly:\\
[4pt]
\textcolor{infoColor}{\texttt{<information>}}\\[-2pt]
\quad [[e\_1]]: info1\\
\quad [[e\_2]]: info2\\
\quad …\\
\textcolor{infoColor}{\texttt{</information>}}\\[-2pt]
- You will be given claim related information above.\\
[8pt]
\textcolor{thinkColor}{\texttt{<think>…</think>}} \\[-2pt]
- Use for every piece of reasoning.\\
- During reasoning, you must verify the claim step by step based on the given information.\\
[8pt]
\textcolor{answerColor}{\texttt{<answer>}}\\[-2pt]
\quad Label: SUPPORT / REFUTE / NOT ENOUGH INFO\\
\quad Evidence: [[e\_1]], [[e\_3]], ...\\
\textcolor{answerColor}{\texttt{</answer>}}\\[-2pt]
- Emit exactly once at the end, no extra text or tags.\\
- Evidence id such as e\_1 will be replaced by real ids from the corpus. You must include useful real ids when answering\\
- Evidence outputs must strictly enforce the format [[e\_i]], [[e\_j]]…\\
- Answer Labels respectively stand for:\\
\quad SUPPORT: The claim is consistent with the cited evidence and the evidence is sufficient to confirm the claim.\\
\quad REFUTE: The claim contradicts the cited evidence and the evidence is sufficient to disprove the claim.\\
\quad NOT ENOUGH INFO: The available evidence is insufficient to determine whether the claim is true or false.\\
[8pt]
Verify the claim:\\[-2pt]
\texttt{\{claim\}}\\
\textcolor{infoColor}{\texttt{<information>}}\\[-2pt]
\texttt{\{evidence\}}\\
\textcolor{infoColor}{\texttt{</information>}}\\
\end{tcolorbox}%
}
\caption{System Prompt for Offline Claim Verification.}
\label{tab:Offline_RL}
\end{figure*}

%% file: tables/veri_acc.tex
\begin{table*}[!t]
\centering
\resizebox{1.0\linewidth}{!}{
\begin{tabular}{l ccc ccc ccc ccc cc}
\toprule
\textbf{Model} 
& \multicolumn{3}{c}{\textbf{FEVEROUS (Veri Acc)}} 
& \multicolumn{3}{c}{\textbf{EX-FEVER (Veri Acc)}} 
& \multicolumn{3}{c}{\textbf{FEVER (Veri Acc)}} 
& \multicolumn{3}{c}{\textbf{SciFACT (Veri Acc)}} 
& \multicolumn{2}{c}{\textbf{HOVER (Veri Acc)}} \\
& \textbf{SUPPORTS} & \textbf{REFUTES} & \textbf{NEI} 
& \textbf{SUPPORT} & \textbf{REFUTE} & \textbf{NEI} 
& \textbf{SUPPORTS} & \textbf{REFUTES} & \textbf{NEI} 
& \textbf{SUPPORT} & \textbf{CONTRADICT} & \textbf{NEI} 
& \textbf{SUPPORTED} & \textbf{NOT SUPPORTED} \\
\midrule
GPT-4o 
& 40.13\% & 38.77\% & 43.19\% 
& 19.77\% & 15.59\% & 44.11\% 
& 70.67\% & 69.33\% & 41.33\% 
& 40.93\% & 43.04\% & 79.32\% 
& 55.80\% & 68.00\% \\
\midrule
Qwen2.5-3B-Instruct 
& 18.37\% & 13.61\% & 40.48\% 
& 12.55\% & 5.32\% & \underline{41.83\%} 
& 48.33\% & 43.00\% & \underline{50.33\%} 
& 26.58\% & 2.53\% & 63.71\% 
& 35.40\% & \underline{65.60\%} \\
Qwen2.5-3B-Instruct-SFT 
& 17.01\% & 13.61\% & \underline{40.82\%} 
& \underline{15.97\%} & 4.56\% & 40.30\% 
& 52.33\% & 40.67\% & 49.67\% 
& 22.78\% & 2.11\% & 67.93\% 
& 33.60\% & 64.60\% \\
Qwen2.5-3B-Instruct-OfflineRL 
& \underline{26.19\%} & 23.81\% & 33.67\% 
& 13.69\% & 6.84\% & 40.68\% 
& 57.33\% & 52.67\% & 43.67\% 
& \textbf{28.69\%} & \underline{10.13\%} & \underline{69.20\%} 
& 39.20\% & 62.80\% \\
Qwen2.5-7B-Instruct 
& 25.17\% & \underline{27.89\%} & 23.47\% 
& 15.21\% & \textbf{10.65\%} & 27.00\% 
& \textbf{61.67\%} & \textbf{63.67\%} & 24.00\% 
& \textbf{28.69\%} & \underline{10.13\%} & 54.01\% 
& \underline{41.00\%} & \textbf{73.20\%} \\
Qwen2.5-3B-Instruct-OnlineRL 
& \textbf{33.67\%} & \textbf{29.93\%} & \textbf{45.24\%} 
& \textbf{19.77\%} & \textbf{10.65\%} & \textbf{66.92\%} 
& \underline{58.00\%} & \underline{54.67\%} & \textbf{55.00\%} 
& 19.41\% & \textbf{14.77\%} & \textbf{90.30\%} 
& \textbf{44.80\%} & 65.40\% \\
\midrule
Llama3.2-3B-Instruct 
& 17.35\% & 4.42\% & \underline{36.05\%} 
& 14.83\% & 3.04\% & 42.59\% 
& 48.00\% & 32.67\% & 32.33\% 
& 25.32\% & 0.42\% & 42.19\% 
& 26.60\% & 46.00\% \\
Llama3.2-3B-Instruct-SFT 
& 17.35\% & 7.14\% & 30.95\% 
& 16.73\% & 3.80\% & 35.74\% 
& 49.67\% & 36.33\% & \textbf{39.00\%} 
& 26.58\% & 0.42\% & 45.99\% 
& 29.60\% & 47.20\% \\
Llama3.2-3B-Instruct-OfflineRL 
& 14.63\% & 15.99\% & 35.03\% 
& 16.35\% & \underline{11.41\%} & 34.60\% 
& 43.33\% & 43.00\% & 27.00\% 
& 12.66\% & 1.27\% & \underline{54.01\%} 
& 30.80\% & 50.40\% \\
Llama3.1-8B-Instruct 
& \underline{23.47\%} & \underline{19.39\%} & \textbf{39.12\%} 
& \underline{22.43\%} & 8.75\% & \underline{49.43\%} 
& \underline{66.67\%} & \textbf{59.00\%} & 33.67\%
& \textbf{37.13\%} & \underline{2.11\%} & 43.04\% 
& \textbf{50.80\%} & \underline{60.40\%} \\
Llama3.2-3B-Instruct-OnlineRL 
& \textbf{26.53\%} & \textbf{20.41\%} & 31.97\% 
& \textbf{25.10\%} & \textbf{13.69\%} & \textbf{52.09\%} 
& \textbf{67.33\%} & \underline{58.00\%} & \underline{34.67\%} 
& \underline{30.80\%} & \textbf{12.66\%} & \textbf{74.26\%} 
& \underline{48.40\%} & \textbf{61.40\%} \\
\bottomrule
\end{tabular}
}
\caption{Evaluation of \textbf{Verification Accuracy} across five datasets. Within each model group, \textbf{bold} denotes the best performance and ~\underline{underline} denotes the second-best.}
\label{tab:veri_results_all}
\end{table*}

%% file: tables/label_acc.tex
\begin{table*}[!t]
\centering
\resizebox{1.0\linewidth}{!}{
\begin{tabular}{l ccc ccc ccc ccc cc}
\toprule
\textbf{Model} 
& \multicolumn{3}{c}{\textbf{FEVEROUS (Label Acc)}} 
& \multicolumn{3}{c}{\textbf{EX-FEVER (Label Acc)}} 
& \multicolumn{3}{c}{\textbf{FEVER (Label Acc)}} 
& \multicolumn{3}{c}{\textbf{SciFACT (Label Acc)}} 
& \multicolumn{2}{c}{\textbf{HOVER (Label Acc)}} \\
& \textbf{SUPPORTS} & \textbf{REFUTES} & \textbf{NEI} 
& \textbf{SUPPORT} & \textbf{REFUTE} & \textbf{NEI} 
& \textbf{SUPPORTS} & \textbf{REFUTES} & \textbf{NEI} 
& \textbf{SUPPORT} & \textbf{CONTRADICT} & \textbf{NEI} 
& \textbf{SUPPORTED} & \textbf{NOT SUPPORTED} \\
\midrule
GPT-4o 
& 86.05\% & 65.64\% & 43.19\% 
& 50.19\% & 69.96\% & 44.11\% 
& 90.33\% & 91.33\% & 41.33\% 
& 69.20\% & 76.37\% & 79.32\% 
& 79.20\% & 68.00\% \\
\midrule
Qwen2.5-3B-Instruct 
& 50.95\% & 35.36\% & \underline{41.83\%} 
& 78.23\% & 29.93\% & 40.48\% 
& 71.00\% & 67.00\% & \underline{50.33\%} 
& 48.52\% & 51.05\% & 63.71\% 
& 59.00\% & \underline{65.60\%} \\
Qwen2.5-3B-Instruct-SFT 
& \underline{51.33\%} & 36.50\% & 40.30\% 
& 71.09\% & 30.27\% & \underline{40.82\%} 
& 73.67\% & 62.67\% & 49.67\% 
& 48.95\% & 49.79\% & 67.93\% 
& 57.20\% & 64.60\% \\
Qwen2.5-3B-Instruct-OfflineRL 
& 50.57\% & 59.32\% & 40.68\% 
& 75.85\% & 50.68\% & 33.67\% 
& \underline{80.00\%} & \underline{78.00\%} & 43.67\% 
& \underline{65.82\%} & \underline{60.34\%} & \underline{69.20\%} 
& \underline{60.20\%} & 62.80\% \\
Qwen2.5-7B-Instruct 
& 44.87\% & \textbf{67.30\%} & 27.00\% 
& \underline{78.91\%} & \textbf{59.52\%} & 23.47\% 
& 78.33\% & 75.33\% & \textbf{55.00\%} 
& \textbf{76.37\%} & \textbf{67.51\%} & 54.01\% 
& 58.20\% & \textbf{73.20\%} \\
Qwen2.5-3B-Instruct-OnlineRL 
& \textbf{52.85\%} & \underline{63.50\%} & \textbf{66.92\%} 
& \textbf{79.25\%} & \underline{59.18\%} & \textbf{45.24\%} 
& \textbf{86.67\%} & \textbf{87.67\%} & 24.00\% 
& 45.57\% & 54.43\% & \textbf{90.30\%} 
& \textbf{62.00\%} & 65.40\% \\
\midrule
Llama3.2-3B-Instruct 
& 47.91\% & 21.29\% & 42.59\% 
& 68.37\% & 19.73\% & \underline{36.05\%} 
& 77.33\% & 51.67\% & 32.33\% 
& 78.06\% & 22.36\% & 42.19\% 
& 66.00\% & 46.00\% \\
Llama3.2-3B-Instruct-SFT 
& 49.43\% & 27.00\% & 35.74\% 
& 74.49\% & 25.51\% & 30.95\% 
& 77.67\% & 51.33\% & \textbf{39.00\%} 
& \underline{80.59\%} & 24.89\% & 45.99\% 
& 67.60\% & 47.20\% \\
Llama3.2-3B-Instruct-OfflineRL 
& 42.59\% & \underline{48.29\%} & 34.60\% 
& 60.54\% & 36.73\% & 35.03\% 
& 80.33\% & 65.67\% & 27.00\% 
& 61.18\% & 40.08\% & \underline{54.01\%} 
& 64.00\% & 50.40\% \\
Llama3.1-8B-Instruct 
& \textbf{60.84\%} & 46.77\% & \underline{49.43\%} 
& \underline{75.85\%} & \underline{37.76\%} & \textbf{39.12\%} 
& \textbf{88.67\%} & \textbf{82.00\%} & \underline{34.67\%} 
& \textbf{88.19\%} & \underline{45.57\%} & 43.04\% 
& \textbf{75.00\%} & \underline{60.40\%} \\
Llama3.2-3B-Instruct-OnlineRL 
& \underline{60.08\%} & \textbf{65.78\%} & \textbf{52.09\%} 
& \textbf{85.03\%} & \textbf{42.52\%} & 31.97\% 
& \underline{88.00\%} & \underline{80.67\%} & 33.67\% 
& 60.34\% & \textbf{64.98\%} & \textbf{74.26\%} 
& \underline{69.40\%} & \textbf{61.40\%} \\
\bottomrule
\end{tabular}
}
\caption{Evaluation of \textbf{Label Accuracy} across five datasets. Within each model group, \textbf{bold} denotes the best performance and ~\underline{underline} denotes the second-best.}
\label{tab:label_results_all}
\end{table*}